%% file: main.tex
\documentclass{article} 
\usepackage{iclr2025_conference,times}
\usepackage{tocloft} 

\setcitestyle{numbers,square}
\makeatletter
\renewcommand\NAT@open{[}
\renewcommand\NAT@close{]}
\renewcommand\NAT@sep{,}
\renewcommand\NAT@citetp{\@firstofone}
\makeatother

\usepackage{url}

\usepackage{lmodern}
\usepackage{pifont}

\usepackage[utf8]{inputenc} 
\usepackage[T1]{fontenc}    
\usepackage{url}            
\usepackage{booktabs}       
\usepackage{amsfonts}       
\usepackage{nicefrac}       
\usepackage{microtype}      
\usepackage{xcolor}         
\usepackage{times}
\usepackage{epsfig}
\usepackage{amssymb}
\usepackage{booktabs}
\usepackage{multirow}
\usepackage{arydshln}
\usepackage{adjustbox}
\usepackage{bm}
\usepackage{nicefrac}
\usepackage{longtable}        
\usepackage{setspace}
\usepackage{hyperref}
\usepackage{wrapfig}
\hypersetup{colorlinks=true, linkcolor=red, citecolor=citecolor, urlcolor=blue}
\definecolor{citecolor}{HTML}{4169E1}
\usepackage{subcaption}
\usepackage{graphicx}

\usepackage{booktabs}
\usepackage{wrapfig}

\usepackage{xspace}
\usepackage{interval}
\usepackage{bm,upgreek}
\usepackage{colortbl}
\usepackage{enumitem}
\usepackage{amssymb}
\usepackage{pifont}
\usepackage{graphicx}
\usepackage{multirow}
\usepackage{amsmath,amssymb}

\usepackage{algorithm}             
\usepackage{algorithmicx}          
\usepackage{algpseudocode}         
\usepackage{float}                 

\usepackage{booktabs}
\usepackage{booktabs}
\usepackage{tikz}

\definecolor{KeywordColor}{RGB}{0,102,153}   
\definecolor{CommentColor}{RGB}{150,150,150} 

\algrenewcommand\algorithmicrequire{\textcolor{KeywordColor}{\textbf{Require:}}}
\algrenewcommand\algorithmicensure{\textcolor{KeywordColor}{\textbf{Ensure:}}}
\algrenewcommand\algorithmicfor{\textcolor{KeywordColor}{\textbf{for}}}
\algrenewcommand\algorithmicend{\textcolor{KeywordColor}{\textbf{end}}}
\algrenewcommand\algorithmicif{\textcolor{KeywordColor}{\textbf{if}}}
\algrenewcommand\algorithmicthen{\textcolor{KeywordColor}{\textbf{then}}}
\algrenewcommand\algorithmicelse{\textcolor{KeywordColor}{\textbf{else}}}
\algrenewcommand\algorithmicreturn{\textcolor{KeywordColor}{\textbf{return}}}
\algrenewcommand\algorithmiccomment[1]{\hfill\textcolor{CommentColor}{\# #1}}

\definecolor{KeywordColor}{RGB}{0,102,153}   
\definecolor{CommentColor}{RGB}{150,150,150} 
\definecolor{StageColor}{RGB}{153,0,153}     

\algrenewcommand\algorithmicrequire{\textcolor{KeywordColor}{\textbf{Require:}}}
\algrenewcommand\algorithmicensure{\textcolor{KeywordColor}{\textbf{Ensure:}}}
\algrenewcommand\algorithmiccomment[1]{\hfill\textcolor{CommentColor}{\# #1}}

\newcolumntype{L}[1]{>{\ttfamily\raggedright\arraybackslash}p{#1}}
\newcolumntype{C}[1]{>{\centering\arraybackslash}m{#1}}

\definecolor{greencheck}{RGB}{0,128,0}
\definecolor{redcross}{RGB}{255,0,0}

\definecolor{gold}{RGB}{255,215,0}
\definecolor{lightgold}{RGB}{255,239,153}

\definecolor{silver}{RGB}{192,192,192}
\definecolor{lightsilver}{RGB}{224,224,224} 

\definecolor{bronze}{RGB}{205,127,50}
\definecolor{lightbronze}{RGB}{233,206,170}

\title{First-Place Solution to NeurIPS 2024 Invisible Watermark Removal Challenge}

\author{Fahad Shamshad$^{1}$, Tameem Bakr$^{1}$, Yahia Shaaban$^{1}$, \\ \textbf{Noor Hussein$^{1,2}$,  
   Karthik Nandakumar$^{1,2}$, Nils Lukas$^{1}$}\\
  $^{1}$Mohamed bin Zayed University of Artificial Intelligence (MBZUAI), UAE\\
  $^{2}$Michigan State University (MSU), USA\\
  \texttt{\{firstname.lastname\}@mbzuai.ac.ae}\\
}

\iclrfinalcopy 
\begin{document}

\maketitle

\begin{abstract}
Content watermarking is an important tool for the authentication and copyright protection of digital media. 
However, it is unclear whether existing watermarks are robust against adversarial attacks. 
We present the \textbf{winning solution} to the NeurIPS 2024 \textit{Erasing the Invisible} challenge, which stress-tests watermark robustness under varying degrees of adversary knowledge. 
The challenge consisted of two tracks: a black-box and beige-box track, depending on whether the adversary knows which watermarking method was used by the provider. 
For the \textbf{beige-box} track, we leverage an \textit{adaptive} VAE-based evasion attack, with a test-time optimization and color-contrast restoration in CIELAB space to preserve the image's quality. 
For the \textbf{black-box} track, we first cluster images based on their artifacts in the spatial or frequency-domain. 
Then, we apply image-to-image diffusion models with controlled noise injection and semantic priors from ChatGPT-generated captions to each cluster with optimized parameter settings. 
Empirical evaluations demonstrate that our method successfully \textbf{achieves near-perfect watermark removal} (95.7\%) with negligible impact on the residual image's quality.
We hope that our attacks inspire the development of more robust image watermarking methods.
\end{abstract}

\section{Introduction}

Content watermarking is a widely used technique for embedding imperceptible information into digital media to ensure provenance verification~\cite{farid2022creating}, copyright protection~\cite{vsarvcevic2024u,xu2024copyrightmeter}, and content traceability~\cite{liu2024survey}.  With generative AI systems now capable of producing high-fidelity synthetic content at scale, watermarking serves as an essential safeguard  for content owners and organizations to combat unauthorized distribution~\cite{zhao2024sok,wang2024security} and forgery~\cite{ren2024sok,aremu2025mitigating}. The goal of a watermarking method is to hide a signal (message) in generated content that can only be detected with a secret watermarking key~\cite{aberna2024digital,lukas2024analyzing}. However, despite its widespread deployment, watermarking systems remain vulnerable to both unintentional distortions (\textit{e.g.,} blur, resizing) and targeted adversarial attacks that intentionally suppress the watermark signal without perceptually degrading the content~\cite{anwaves,diaa2024optimizing}.

A robust watermarking scheme must ensure that successful removal is only possible at the cost of significant perceptual degradation~\cite{zhao2024sok,lukas2023ptw,holtervennhoff2025security}. However, recent studies have exposed critical vulnerabilities in existing methods, where adversaries can remove or spoof watermark signals by exploiting adaptive attacks~\cite{lukas2024leveraging,diaa2024optimizing}, latent-space priors~\cite{muller2025black,jain2025forging}, or even simple averaging techniques~\cite{yang2024can}. These attacks preserve visual fidelity, often introducing no perceptible artifacts, thereby undermining the reliability of watermark detectors. As watermarking becomes increasingly important for ensuring the integrity, attribution, and traceability of AI-generated content, addressing these weaknesses is imperative. Progress in this area requires rigorous evaluation under realistic threat models to inform the design of watermarking schemes that are not only imperceptible but also resilient to both adversarial manipulations and common distortions.

To gauge the robustness of invisible watermarking methods against realistic threats, the NeurIPS 2024 \textit{Erasing the Invisible: A Stress-Test Challenge for Image Watermarks}~\cite{ding2024erasing} introduced a rigorous benchmark targeting watermark removal. This paper presents our \textbf{first-place winning solution} to this challenge under two practical threat models: \textbf{beige-box}, where the watermarking methodology was known, and \textbf{black-box}, where no prior knowledge was available. For the beige-box scenario, we design an adaptive VAE-based attack for effective watermark removal, coupled with test-time optimization and frequency-aware color restoration to preserve image quality. We also show that simple spatial-domain translations can effectively disrupt TreeRing-based watermarks. In the black-box setting, we design targeted, cluster-specific removal attacks by grouping images based on spatial and spectral artifacts, followed by diffusion-based purification guided by semantic captions. \textit{Our approach outperformed the runner-ups by 26\% and 31.7\% in detection score on the beige-box and black-box tracks respectively, while preserving high perceptual quality (see Table~\ref{table:black-box_track} and Table~\ref{table:beige-box_track})}. By exposing vulnerabilities in existing watermarking methods, we aim to inspire the development of more robust defenses against such attacks.

\begin{table}[t]
\centering
\begin{minipage}{0.495\textwidth}
\centering
\caption{Black-box Track Final Leaderboard}
\label{table:black-box_track}
\resizebox{\textwidth}{!}{
\begin{tabular}{ccccc}
\toprule
\multicolumn{5}{c}{\textbf{Black-box Track}} \\ 
\midrule
Rank & Participant       & Detection & Quality & Total \\
\midrule
\rowcolor{gold!15}
\raisebox{-0.2\height}{\begin{tikzpicture}
\node[draw=black, fill=gold, circle, text=black, inner sep=0.2pt] {1};
\end{tikzpicture}} & \textbf{Team-MBZUAI}      & \textbf{0.043}         & \textbf{0.136}                           & \textbf{0.143}                    \\
\rowcolor{silver!10}
\raisebox{-0.2\height}{\begin{tikzpicture}
\node[draw=black, fill=silver, circle, text=black, inner sep=0.2pt] {2};
\end{tikzpicture}} & Team-SHARIF   & 0.063        & 0.158                           & 0.170                    \\
\rowcolor{bronze!10}
\raisebox{-0.2\height}{\begin{tikzpicture}
\node[draw=black, fill=bronze, circle, text=black, inner sep=0.2pt] {3};
\end{tikzpicture}} & Team-UFL        & 0.087        & 0.177                           & 0.197                    \\ 
\bottomrule
\end{tabular}
}
\end{minipage}
\hfill
\begin{minipage}{0.495\textwidth}
\centering
\caption{Beige-box Track Final Leaderboard}
\label{table:beige-box_track}
\resizebox{\textwidth}{!}{
\begin{tabular}{ccccc}
\toprule
\multicolumn{5}{c}{\textbf{Beige-box Track}} \\ 
\midrule
Rank & Participant       & Detection & Quality & Total  \\
\midrule
\rowcolor{gold!15}
\raisebox{-0.2\height}{\begin{tikzpicture}
\node[draw=black, fill=gold, circle, text=black, inner sep=0.2pt] {1};
\end{tikzpicture}} & \textbf{Team-MBZUAI}      & \textbf{0.037}         & \textbf{0.153}                           & \textbf{0.157}                    \\
\rowcolor{silver!10}
\raisebox{-0.2\height}{\begin{tikzpicture}
\node[draw=black, fill=silver, circle, text=black, inner sep=0.2pt] {2};
\end{tikzpicture}} & Team-SONY   & 0.050        & 0.176                           & 0.183                    \\
\rowcolor{bronze!10}
\raisebox{-0.2\height}{\begin{tikzpicture}
\node[draw=black, fill=bronze, circle, text=black, inner sep=0.2pt] {3};
\end{tikzpicture}} & Team-SHARIF        & 0.127        & 0.222                           & 0.256                    \\
\bottomrule
\end{tabular}
}
\end{minipage}
\end{table}

\vspace{-0.5em}
\begin{figure*}[t]
   \centering
   \includegraphics[width=0.95\textwidth]{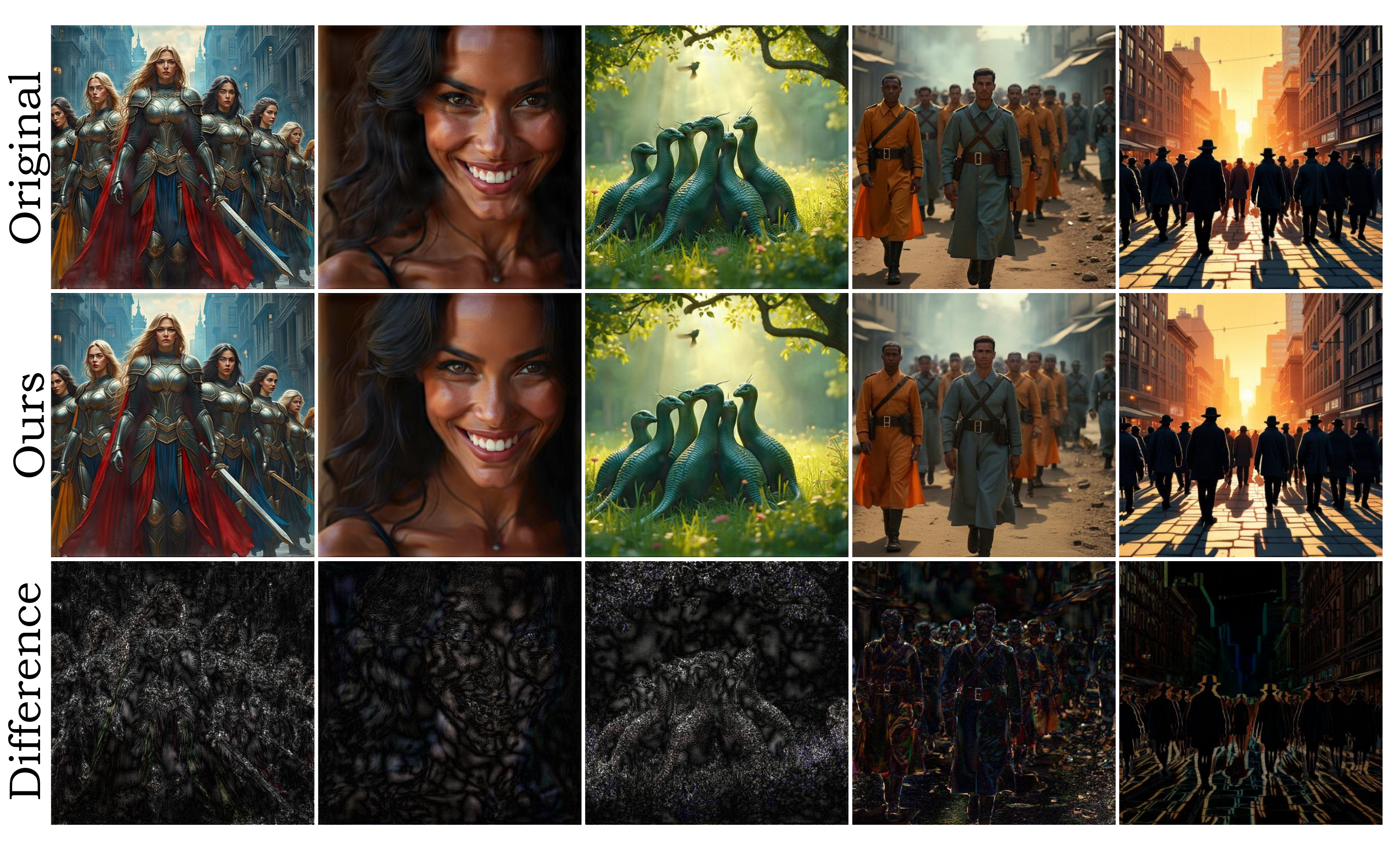} 
   \vspace{-1em}
   \caption{\textit{Top row}: Original watermarked images. \textit{Bottom row}: Images after our attack, with minimal perceptual difference from the originals, showcasing the effectiveness of our method in preserving visual fidelity. Best viewed zoomed in.}
\end{figure*}

\section{Related Work}
\textbf{Image Watermarking.} Image watermarking embeds imperceptible signals into digital images to support authentication, copyright enforcement, and forensic traceability~\cite{chen2025image,qi2022overview,potdar2005survey,duan2025visual}. Traditional image watermarking approaches operate in the spatial or frequency domains by modifying pixel values or transform-domain coefficients such as Discrete Cosine Transform~\cite{bors1996image}, Discrete Wavelet Transforms~\cite{al2007combined}, or Discrete Fourier Transform~\cite{kang2003dwt}, often trading off between imperceptibility and robustness. More recently, deep learning-based watermarking schemes leverage convolutional neural networks and generative models to embed watermarks via learned feature spaces~\cite{zhang2018protecting,hosny2024digital}. Methods such as StegaStamp~\cite{tancik2020stegastamp}, Gaussian Shading~\cite{yang2024gaussian}, IConMark~\cite{sadasivan2025iconmark}, Robin~\cite{huang2024robin}, and TreeRing~\cite{wen2024tree} improve robustness under common corruptions (\textit{e.g.,} JPEG compression, resizing) and are increasingly adopted to tag AI-generated content. Despite their effectiveness in benign settings, these techniques remain vulnerable to both incidental degradation and targeted removal attacks~\cite{anwaves,zhao2024invisible}, underscoring the need for watermarking methods with stronger resilience.

\textbf{Robustness of Image Watermarks.} 
Image watermarks are vulnerable to degradation from common distortions such as Gaussian noise, blurring, and compression. More critically, adversarial attacks intentionally exploit model vulnerabilities to remove the watermark signal while maintaining high perceptual quality~\cite{hwang2024invisible,yang2024can,gluch2024good}. 
To mitigate these threats, recent approaches employ adversarial training, where watermarking models are optimized to withstand a range of perturbations~\cite{huang2024robin,thakkar2023elevating}. Generative models, including autoencoders and diffusion models, have also been leveraged to embed more resilient watermarks by aligning with the natural image manifold~\cite{huang2024robin}. Despite these advances, achieving robustness without compromising imperceptibility remains an open challenge, especially with partial or complete knowledge of the watermarking algorithm~\cite{ma2025safety,fairoze2025difficulty}. The NeurIPS 2024 \textit{Erasing the Invisible challenge}~\cite{ding2024erasing} provides a benchmark for evaluating the resilience of state-of-the-art image watermarking methods under beige-box and black-box threat models.

\textbf{Generative Models.}
Pretrained generative models have demonstrated strong performance across a range of vision tasks, including image restoration~\cite{asim2020blind,zhao2023generative,shamshad2020compressed,shamshad2019subsampled,shamshad2019deep,shamshad2019adaptive}, privacy preservation~\cite{shamshad2023clip2protect,shamshad2023evading}, and adversarial purification~\cite{nie2022diffusion,samangouei2018defense}. Leveraging their powerful priors over natural images, models such as VAEs~\cite{kingma2013auto}, GANs~\cite{goodfellow2020generative}, and diffusion models~\cite{ho2020denoising} can reconstruct high-fidelity content from corrupted or perturbed inputs. These capabilities have recently been explored in the context of watermarking, where the goal is to remove imperceptible signals embedded within images while preserving visual quality~\cite{liu2ssss024image,zhao2024invisible}. For example, VAEs can project watermarked images into clean latent spaces that suppress hidden signals, while diffusion-based regeneration/rinsing methods have shown promise in erasing invisible perturbations~\cite{liu2024image,anwaves}. Recently, inspired by the inductive bias of untrained neural networks~\cite{ulyanov2018deep,qayyum2022untrained,qayyum2020single}, such architectures have also been explored for invisible image watermark removal~\cite{liang2025baseline}, demonstrating surprising effectiveness.

\section{Proposed Approach}

\textbf{Challenge Overview:} The NeurIPS 2024 competition \textit{Erasing the Invisible: A Stress-Test Challenge for Image Watermarks~\cite{ding2024erasing}} rigorously assesses invisible watermarking robustness under realistic threat models.
The benchmark comprises \textbf{300 watermarked images per track} across two settings: a \textbf{beige-box} scenario, where the attacker knows the watermarking algorithm but not its parameters, and a \textbf{black-box} scenario, where no prior knowledge is available. Participants were tasked with removing invisible watermarks while preserving perceptual image quality. Evaluation was based on two criteria:

\begin{itemize}
    \item \textbf{Detection score} measures watermark removal success as the true positive rate (TPR) at a 0.1\% false positive rate (FPR). For each method, a detection threshold is set using the 0.001st percentile of distances \( d(m, m') \) between known messages \( m \) and decoded outputs \( m' \) from 10{,}000 unwatermarked images. An attacked image is flagged as watermarked if its decoded message falls below this threshold. The detection score is then the fraction of images that were originally watermarked still flagged as containing a watermark.

    \item \textbf{Image quality score} is a weighted combination of low-level fidelity metrics (PSNR, SSIM~\cite{wang2004image}, NMI) and high-level perceptual metrics (FID, CLIP-FID~\cite{kynkaanniemi2022role}, LPIPS~\cite{zhang2018unreasonable}, Delta Aesthetics~\cite{ding2024erasing}, and Delta Artifacts~\cite{ding2024erasing}). Each metric is normalized based on attack sensitivity and weighted accordingly: positive weights for metrics where lower is better (\textit{e.g.}, FID), and negative for those where higher is better (\textit{e.g.}, SSIM). This comprehensive multi-metric formulation ensures consistency and fairness across both pixel-level and perceptual domains.
\end{itemize}

\textit{We can query the competition leaderboard up to five times per day for both beige-box and black-box tracks to evaluate our submission performance}. Below, we provide details of our proposed attack for both the beige-box and black-box tracks.

\subsection{\textbf{Beige-Box Track}}

\begin{figure}[t]
    \centering
    \includegraphics[clip, trim=20 50 10 15,width=1\textwidth]{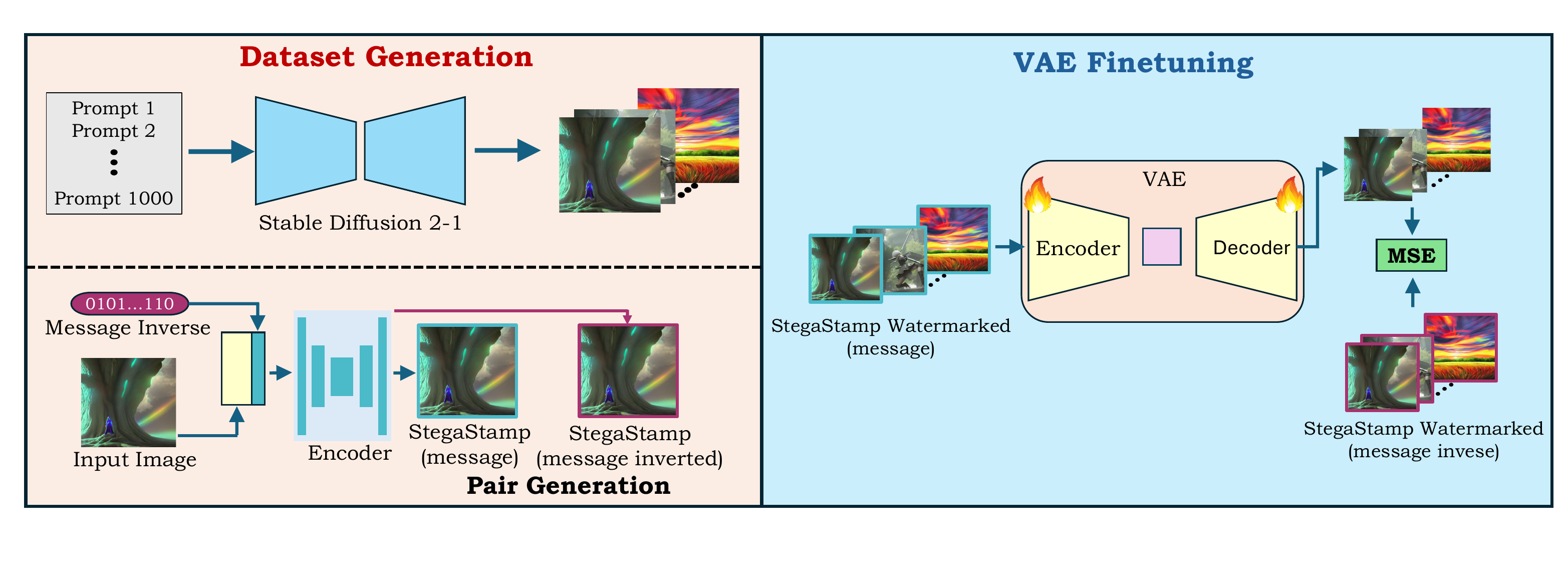} 
    \caption{\small \textbf{Overview of our dataset generation and VAE-based watermark removal pipeline for StegaStamp watermarks}. We generate 1,000 images from Stable Diffusion 2-1 using publicly available prompts, embedding each with a random binary message and its inverse via a StegaStamp encoder to create paired watermarked images. A fine-tuned Variational Autoencoder (VAE) minimizes Mean Squared Error (MSE) loss between pairs, effectively suppressing watermark artifacts while preserving perceptual quality.}
    \label{fig:example}
\end{figure}

\begin{wraptable}{r}{0.5\textwidth}
    \vspace{-1em}
    \centering
    \small
    \setlength{\tabcolsep}{5pt} 
    \caption{\small Performance comparison of our watermark removal pipeline. \textbf{Det.} refers to the detection metric (lower is better), \textbf{Qual.} measures image quality. Progressive enhancements through test-time optimization and color/contrast adjustment improve quality.}
    \label{tab:results_}
    \begin{tabular}{lccc}
        \toprule
        \rowcolor{blue!20}\textbf{Method} & \textbf{Det.} & \textbf{Qual.} & \textbf{Overall} \\
        \midrule
        \rowcolor{blue!5} VAE Finetune & \textbf{0.023} & 0.192 & 0.193 \\
        \rowcolor{blue!10} + Test-Time Opt. & 0.033 & 0.161 & 0.165 \\
        \rowcolor{blue!15} + Color/Contrast & 0.037 & \textbf{0.153} & \textbf{0.157} \\
        \bottomrule
    \end{tabular}
\end{wraptable}

In the beige-box track, we had access to the algorithmic description of the watermarking method used to embed invisible watermarks in the provided images, but not its hyperparameters or the underlying generative model. This partial prior knowledge enabled the design of an adaptive attack strategy that targeted the known watermarking scheme, akin to recent efforts~\cite{lukas2024leveraging,diaa2024optimizing}. The organizers released 300 watermarked images, equally divided between two watermarking methods: a modified StegaStamp algorithm~\cite{tancik2020stegastamp} and a variant of the TreeRing method~\cite{wen2024tree}. We designed a dedicated watermark removal pipeline for each method.

\subsubsection{Stegastamp Watermark Removal} \label{sec:stegastamp}
For the StegaStamp-based watermark removal, we developed a three-stage pipeline: (\textbf{i}) \textit{paired data generation}, (\textbf{ii}) \textit{adaptive VAE fine-tuning}, and (\textbf{iii}) \textit{post-processing for quality restoration}. This structured methodology enabled effective suppression of the embedded watermark while preserving the perceptual quality of the images. The overall pipeline is shown in Figure~\ref{fig:example}.

\underline{\textit{\textbf{Paired Dataset Generation:}}} We first curated a comprehensive training dataset leveraging 1{,}000 text prompts from the Hugging Face Stable-Diffusion-Prompts dataset~\cite{huggingface2024prompts}. Using these prompts, we generated corresponding images via Stable Diffusion 2-1 with a guidance scale of 7.5 and 50 inference steps. Each generated $512^{2}$ image was resized to $400^2$ pixels using bilinear interpolation before being processed through a pretrained StegaStamp model from the WAVES repository~\cite{anwaves}. The key aspect of our dataset preparation involved creating image pairs where each original image was encoded with both a 100-bit binary message $m$ sampled uniformly at random and its inverse $1-m$, resulting in a dataset of 1,000 paired examples that captured watermarking artifacts.

\begin{algorithm}[t]
\caption{{\color{purple}\textbf{VAE Finetuning}}}
\label{alg:vae-finetune-psi}
\begin{algorithmic}[1]
  \Require Pretrained VAE $(E_\theta,D_\phi)$, dataset $\{(x_w,x_i)\}_{n=1}^N$, 
           learning rate $\alpha$, batch size $B$, epochs $E$
  \State Initialize $\psi \gets \{\theta,\phi\}$ \Comment{Initialize encoder and decoder parameters}
  \For{$\mathrm{epoch}=1$ \textbf{to} $E$}
    \State \textbf{Shuffle} dataset
    \For{each batch $\{(x_w,x_i)\}_{k=1}^B$}
      \State $z \gets E_\theta(x_w)$ \Comment{Encode watermarked image}
      \State $\hat{x} \gets D_\phi(z)$ \Comment{Decode latent representation}
      \State $L \gets \tfrac{1}{B}\sum_{k=1}^B \|\hat{x}^{(k)} - x_i^{(k)}\|_2^2$ \Comment{MSE reconstruction loss}
      \State Compute gradient $\nabla_\psi L$ 
      \State $\psi \gets \psi - \alpha\,\nabla_\psi L$\Comment{Gradient Update via Backpropagation}
    \EndFor
  \EndFor
  \Ensure Finetuned VAE $(E_\theta^*,D_\phi^*)$ \Comment{Return trained VAE parameters}
\end{algorithmic}
\end{algorithm}

\underline{\textbf{Adaptive VAE Finetuning:}} The core of our attack framework centers on a Variational Autoencoder (VAE) that was adaptively tuned to perform watermark removal via supervised reconstruction. Given a watermarked image $x_w$ and its inverse message counterpart $x_i$, the VAE - consisting of encoder $E_\theta$ and decoder $D_\phi$-was optimized to reconstruct $x_i$ from $x_w$, using the following MSE loss:

\begin{equation}
    \mathcal{L}(\theta,\phi) = \|D_\phi(E_\theta(x_w)) - x_i\|^2_2,
\end{equation}

where $D_\phi(E_\theta(x_w))$ represents the reconstructed image from the watermarked input, and $x_i$ is the target image containing the inverted message. We optimize this objective using Adam optimizer with learning rate $\alpha = 1\times10^{-5}$ for 10 epochs with a batch size of 16. To stabilize training, we employ gradient clipping with a maximum norm of 1.0. Model training was performed on an NVIDIA A6000 GPU (48GB VRAM) and completed in under two GPU hours. We adapted a pretrained SDXL VAE model. This adaptive fine-tuning stage enabled the model to effectively strip away the watermark while preserving image structure. The VAE Finetuning algorithm is given in Algorithm.~\ref{alg:vae-finetune-psi}.

\begin{figure*}[t]
   \centering
   \includegraphics[clip, trim=27 60 130 40, width=1\textwidth]{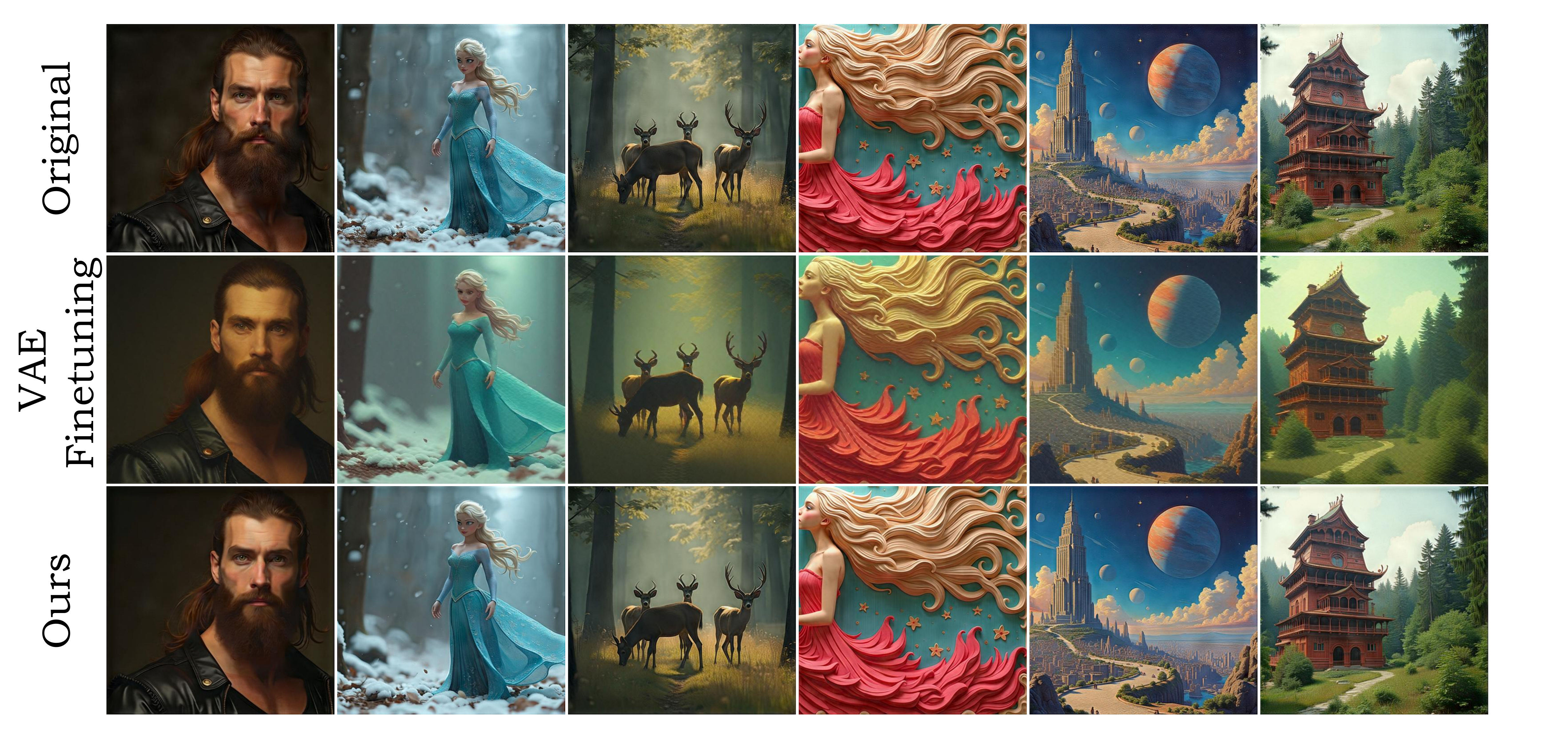} 
   \vspace{-0em}
   \caption{\textit{Top row}: Original watermarked images. \textit{Bottom row}: Images after our attack, with minimal perceptual difference from the originals, showcasing the effectiveness of our method in preserving visual fidelity. Best viewed zoomed in.}
   \label{fig:sample_figure} 
\end{figure*}

\underline{\textbf{Quality-Preserving Post-Processing:}} Despite strong watermark removal performance, VAE reconstructions degraded color and contrast fidelity as shown in Figure.~\ref{fig:sample_figure}. To address this issue, we introduced a two-stage post-processing pipeline (Algorithm~\ref{alg:vae-inference}) with the aim to enhance the image quality without re-introducing the removed watermark. 
\begin{itemize}
    \item \textbf{Test-Time VAE Optimization:} Using the SDXL Refiner VAE, we performed image-specific optimization by fine-tuning the VAE parameters $\{\theta, \phi\}$ to better align with the original watermarked input $x_w$. The loss function combined pixel-wise, perceptual (LPIPS), and structural (SSIM) terms:
    \begin{equation}
    \small
    \mathcal{L}_{\text{total}} = \underbrace{\|D_\phi(E_\theta(x_r)) - x_w\|^2}_\text{MSE Loss} + \underbrace{\mathcal{L}_{\text{LPIPS}}(D_\phi(E_\theta(x_r)), x_w)}_\text{Perceptual Loss} + \underbrace{0.5(1 - \text{SSIM}(D_\phi(E_\theta(x_r)), x_w))}_\text{Structural Similarity Loss}
    \end{equation}
    While this step effectively removes the watermark signal, it can introduce slight degradation in visual quality, motivating the need for the second stage.

\item \textbf{Color and Contrast Transfer:} The second stage restores perceptual quality by adjusting color and contrast in the CIELAB color space. Let \( x_{\text{opt}} \) denote the output from test-time optimization, with CIELAB components \(\{L_{\text{opt}}, a_{\text{opt}}, b_{\text{opt}}\}\), and let \(\{L_w, a_w, b_w\}\) be those of the original watermarked image \( x_w \). 
For \textbf{color transfer}, we preserve the luminance \( L_{\text{opt}} \) from the optimized image and adopt the chrominance components from the watermarked image, yielding an intermediate image \( x_c = \mathcal{F}_{\text{RGB}}(L_{\text{opt}}, a_w, b_w) \), where \( \mathcal{F}_{\text{RGB}} \) denotes conversion from CIELAB to RGB space.
Next, for \textbf{contrast transfer}, we match the statistical moments of the luminance channel. Let \( \mu_c, \sigma_c \) and \( \mu_w, \sigma_w \) represent the mean and standard deviation of the luminance channels of \( x_c \) and \( x_w \), respectively. The adjusted luminance is computed as \( L_{\text{final}} = \frac{\sigma_w}{\sigma_c}(L_c - \mu_c) + \mu_w \).
The final image is reconstructed as:
\[
x_{\text{final}} = \mathcal{F}_{\text{RGB}}(L_{\text{final}}, a_w, b_w).
\]
The color and contrast transfer step enhances visual fidelity without reintroducing the watermark. For an ablation of this stage, see Table~\ref{tab:results_}.

\end{itemize}

As shown in Table \ref{tab:results}, the complete quality-preserving post-processing pipeline significantly outperforms VAE fine-tuning alone across all evaluation metrics. Specifically, PSNR improves by over 6 dB and SSIM increases by 0.176, while perceptual distance measures such as LPIPS, FID, and CLIPFID are markedly reduced. These gains highlight the effectiveness of combining image-specific test-time optimization with CIELAB-based color and contrast transfer, which restores visual fidelity without reintroducing the removed watermark.

\begin{algorithm}[t]
\caption{{\color{purple}\textbf{Test-Time Image Optimization}}}
\label{alg:vae-inference}
\begin{algorithmic}[1]
  \Require Finetuned VAE $(E_{\theta^*},D_{\phi^*})$, watermarked image $x_w$, 
           \textbf{optional} refinement steps $T$, step size $\eta$
  \State \textcolor{StageColor}{\texttt{\#Inference}}
  \State \quad $z \gets E_{\theta^*}(x_w)$  \Comment{Encode watermarked image}
  \State \quad $x_r^{(0)} \gets D_{\phi^*}(z)$ \Comment{Decode latent representation} 
  \If{$T$ \textbf{defined}}
  \State \textcolor{StageColor}{\texttt{\#Test-Time Image Optimization}}
    \State Initialize refiner VAE params $(\theta_r,\phi_r)$
    \For{$t=1$ \textbf{to} $T$}
      \State
        $L_{\mathrm{total}}(\theta_r,\phi_r)
        = \bigl\|D_{\phi_r}(E_{\theta_r}(x_r^{(t-1)})) - x_w\bigr\|_2^2
        + \mathrm{LPIPS} + 0.5\,(1-\mathrm{SSIM})$
      \State $(\theta_r,\phi_r)\;\leftarrow\;(\theta_r,\phi_r)\;-\;\eta\,\nabla_{(\theta_r,\phi_r)}\,L_{\mathrm{total}}$\Comment{Gradient Update}
      \State $x_r^{(t)} \gets D_{\phi_r}(E_{\theta_r}(x_r^{(t-1)}))$
    \EndFor
    \State \textcolor{StageColor}{\texttt{\hspace{-0.0em}\#Color–Contrast Transfer}}
    \Statex \quad \hspace{0.36em} Convert $x_r^{(T)},\,x_w$ to CIELAB: $(L_c,a_c,b_c),(L_w,a_w,b_w)$  
    \Statex \quad \hspace{0.36em} $L_c' \gets \tfrac{\sigma_w}{\sigma_c}(L_c - \mu_c) + \mu_w$  
    \Statex \quad \hspace{0.36em} $x_{\mathrm{final}} \gets \mathrm{CIELAB}\!\to\!\mathrm{RGB}(L_c',\,a_w,\,b_w)$
  \Else
    \State \quad $x_{\mathrm{final}} \gets x_r^{(0)}$
  \EndIf
  \Ensure Recovered image $x_{\mathrm{final}}$
\end{algorithmic}
\end{algorithm}

\begin{table}[t]
\centering
\caption{Quantitative comparison of our proposed quality-preserving watermark removal pipeline against VAE fine-tuning alone. Higher PSNR and SSIM, and lower LPIPS, FID, NMI, and CLIPFID indicate better perceptual quality and watermark suppression. The results demonstrate that integrating test-time optimization with color/contrast transfer yields substantial gains across both pixel-level and perceptual metrics.}
\small
\setlength{\tabcolsep}{6pt}
\begin{tabular}{lcccccc}
    \toprule
    \rowcolor{blue!20}\textbf{Method} & \textbf{PSNR}$\uparrow$ & \textbf{SSIM}$\uparrow$ & \textbf{LPIPS}$\downarrow$ & \textbf{FID}$\downarrow$ & \textbf{NMI}$\uparrow$ & \textbf{CLIPFID}$\downarrow$ \\
    \midrule
    \rowcolor{blue!5} VAE Finetune & 21.899 & 0.647 & 0.264 & 85.383 & 0.231 & 11.805 \\
    \rowcolor{blue!10} \textbf{Ours} & \textbf{28.061} & \textbf{0.823} & \textbf{0.078} & \textbf{30.786} & \textbf{0.326} & \textbf{2.831} \\
    \bottomrule
\end{tabular}
\label{tab:results}
\end{table}

\subsubsection{TreeRing Watermark} \label{sec:treering}
For images embedded with TreeRing watermarks~\cite{wen2024tree}, we discovered a notable vulnerability to phase perturbations in the frequency domain. Specifically, modifying the phase component of the Fourier spectrum induces a spatial translation in the image domain—an operation that can effectively disrupt the watermark signal without introducing noticeable perceptual changes.
\begin{wrapfigure}{r}{0.4\textwidth} 
  \centering
  \includegraphics[clip, trim=63 70 85 50, width=0.4\textwidth]{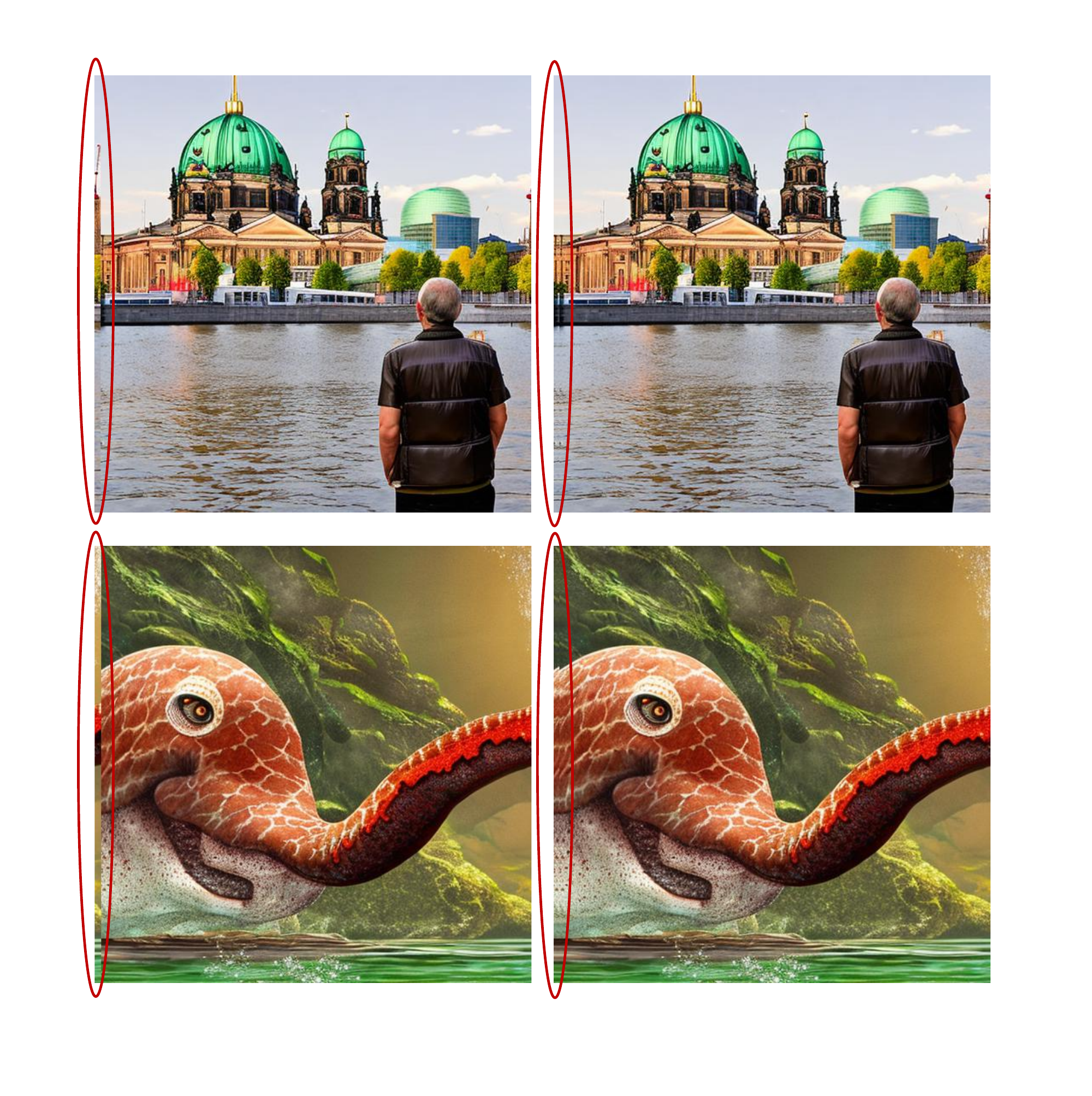}
  \caption{Effect of spatial translation on TreeRing watermarks. \textbf{\textit{Left}}: Direct 7-pixel shift removes watermark but introduces boundary artifacts (highlighted in red). \textbf{\textit{Right}}: Restoring the leftmost 7 columns from the original image removes oundary artifacts while preserving quality.}
  \label{fig:treering-fig}
  \vspace{-4em}
\end{wrapfigure}
Building on this insight, we implemented a lightweight spatial-domain defense by applying a horizontal shift to each image as
$x_{\text{shifted}} = \mathcal{T}(x_w, \Delta x)$, 
where \( x_w \) is the input watermarked image, \( \mathcal{T} \) denotes a horizontal translation operator, and \( \Delta x = 7 \) pixels is the empirically determined optimal shift that achieves a favorable trade-off between watermark removal and perceptual fidelity (see Figure~\ref{fig:treering-fig}).

However, spatial translation alone introduce visual artifacts near image boundaries. To counter this, we selectively restore the leftmost \( \Delta x \) columns from the original image:
\[
x_{\text{final}}(i,j) = \begin{cases} 
x_w(i,j) & \text{if } j < \Delta x \\
x_{\text{shifted}}(i,j) & \text{otherwise}
\end{cases}
\]
This simple yet effective approach removes TreeRing watermarks while maintaining high image quality and requires no training or additional models, making it computationally efficient.

\subsection{\textbf{Black-Box Track}}

\begin{figure*}[t]
  \centering
  \begin{minipage}[t]{0.6\textwidth} 
    \vspace{0pt} 
    \includegraphics[clip, trim=95 660 75 95, width=\textwidth]{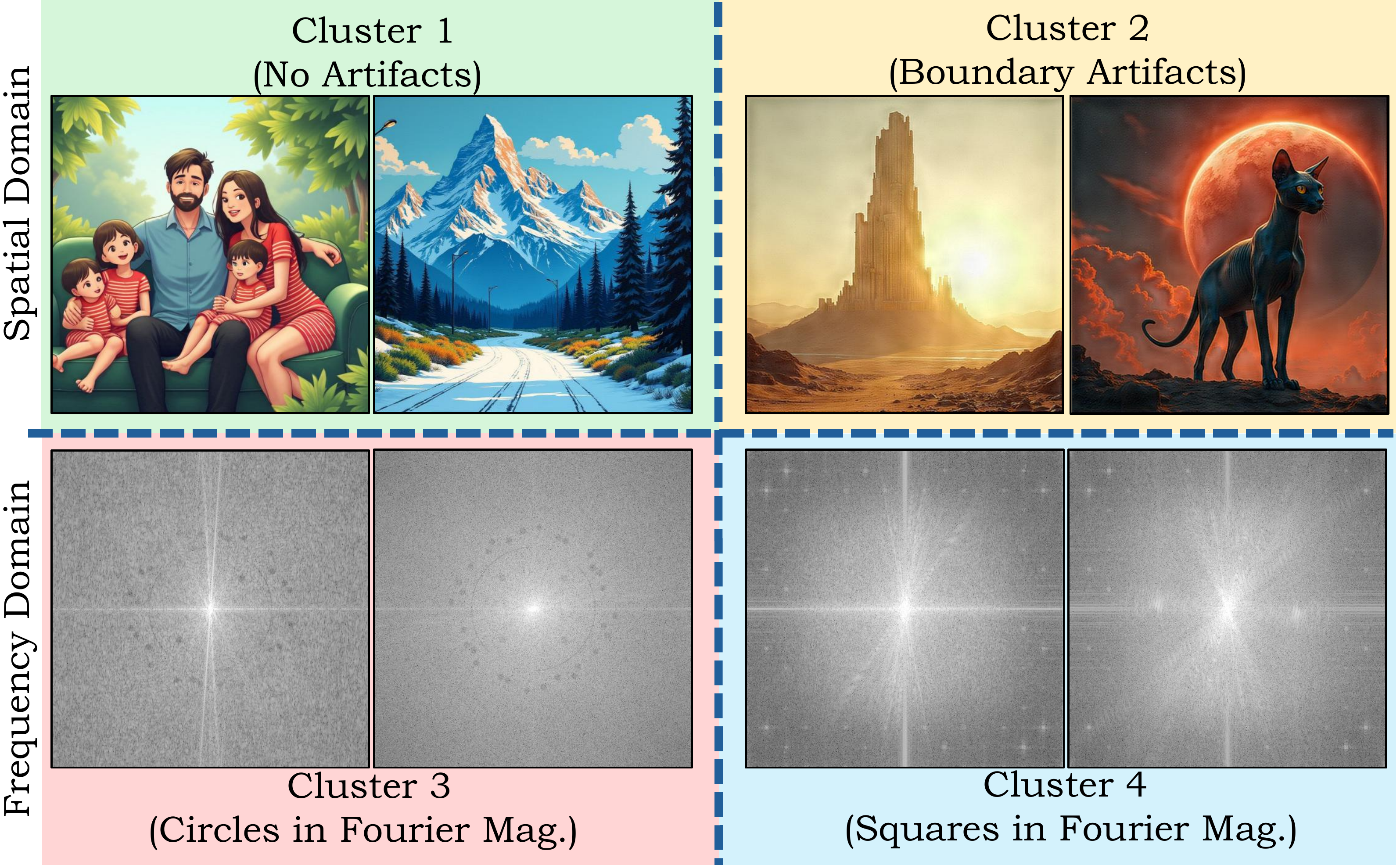}
  \end{minipage}%
  \hfill
  \begin{minipage}[t]{0.39\textwidth} 
    \vspace{0pt} 
    \caption{\small \textbf{Spatial–frequency artifact clustering of 300 black-box watermarked images}. Each image was manually examined for visible spatial and frequency-domain patterns. This yielded four clusters: {\color{teal}\textbf{Cluster-1}} no noticeable artifacts, {\color{olive}\textbf{Cluster-2}} boundary artifacts in the spatial domain, {\color{orange}\textbf{Cluster-3}} circular patterns in the Fourier magnitude spectrum, and {\color{blue}\textbf{Cluster-4}} square patterns in the Fourier magnitude spectrum. Identifying these patterns allowed us to design targeted removal pipelines optimized for the specific artifact type.}
    \label{fig:cluster_}
    \vspace{-1em}
  \end{minipage}
\end{figure*}

In the black-box track, no prior information was provided about the underlying watermarking algorithm. 
We adopted a data-driven strategy to infer potential watermarking mechanisms by systematically analyzing signatures in the spatial and frequency domains. 
Since different watermarking schemes inherently leave characteristic artifact patterns, either as spatial distortions or as structured frequency components, these observations directly guided our clustering process. 
Through comprehensive analysis, we partitioned the 300 watermarked images into four distinct clusters, as shown in Figure~\ref{fig:cluster_}: {\color{teal}\textbf{Cluster-1}}: no discernible artifacts, 
{\color{olive}\textbf{Cluster-2}}: border-like artifacts in the spatial domain, {\color{orange}\textbf{Cluster-3}}: circular patterns in the Fourier magnitude spectrum, {\color{blue}\textbf{Cluster-4}}: square patterns in the Fourier magnitude spectrum.  This clustering allowed us to develop targeted removal strategies tailored to each artifact signature, significantly improving both the efficacy of watermark removal and the preservation of image quality compared to uniform parameter settings.

\begin{figure*}[t]
    \centering
    \vspace{-2pt} 
    \includegraphics[clip, trim=30 50 22 40, width=0.8\textwidth]{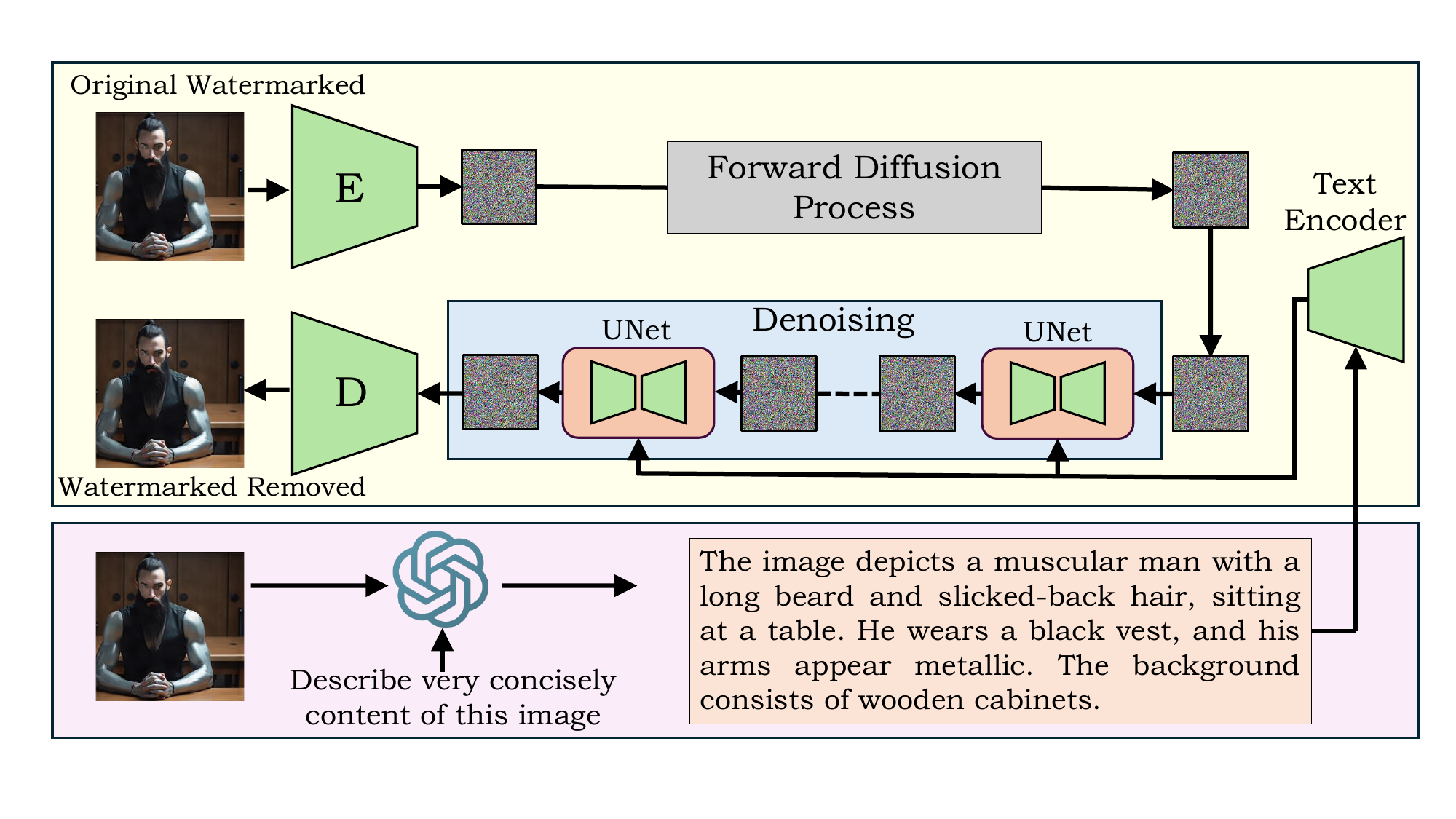}
    \caption{\small \textbf{Overview of our image‑to‑image diffusion pipeline for black‑box watermark removal}. Starting from the watermarked input, we perform a forward diffusion step, injecting Gaussian noise according to a strength parameter $s$ (tuned per cluster), to obtain a noisy latent. A pretrained UNet denoiser, conditioned on semantic embeddings of ChatGPT‑generated captions, then carries out reverse diffusion over $T=500$ steps with guidance scale of 1, gradually reconstructing a high‑fidelity, watermark‑free output. By selecting $s$ per cluster, the pipeline balances aggressive artifact suppression with content preservation.}
    \label{fig:img2img}
    \vspace{-7pt} 
\end{figure*}

\begin{table*}[ht]
  \centering
\caption{\textbf{Cluster assignments for all 300 black-box images}, grouped by manually identified artifact type: boundary artifacts, circular or square Fourier magnitude patterns, or no visible artifacts. Each block lists the image indices belonging to that cluster, enabling targeted strategy design  and facilitating quantitative comparisons of removal performance.}
  \label{tab:cluster-assignments-flat}
  \footnotesize
  \begin{tabular}{>{\centering\arraybackslash}p{0.86\textwidth}} 
    \toprule
     \rowcolor{green!10} \texttt{\textbf{Cluster 1 (No Artifacts)}} \\
    \rowcolor{green!10} \ttfamily{0, 2, 3, 4, 10, 13, 14, 16, 19, 21, 24, 29, 30, 33, 36, 40, 41, 43, 50, 51, 56, 58, 60, 61, 62, 67, 71, 75, 77, 80, 83, 86, 93, 94, 95, 96, 97, 98, 104, 107, 109, 110, 112, 113, 114, 123, 125, 126, 130, 137, 138, 146, 149, 154, 155, 157, 159, 161, 162, 166, 167, 171, 175, 178, 179, 187, 196, 198, 199, 200, 201, 205, 209, 210, 215, 216, 217, 218, 220, 226, 227, 233, 235, 241, 245, 247, 249, 250, 254, 260, 262, 266, 268, 275, 279, 280, 281, 288, 291, 292, 294, 296} \\ \midrule
     \rowcolor{yellow!10} \texttt{\textbf{Cluster 2 (Boundary Artifacts)}} \\ 
    \rowcolor{yellow!10} \ttfamily{1, 6, 11, 23, 35, 42, 48, 49, 57, 65, 68, 69, 78, 82, 84, 85, 91, 100, 102, 105, 119, 120, 121, 131, 132, 140, 141, 142, 143, 148, 169, 170, 180, 186, 203, 204, 207, 231, 240, 246, 256, 257, 261, 267, 269, 274, 283, 297, 298} \\ \midrule
    \rowcolor{orange!10} \texttt{\textbf{Cluster 3 (Circular Patterns in Fourier Magnitude)}} \\
     \rowcolor{orange!10} \ttfamily{9, 17, 25, 28, 31, 32, 38, 39, 47, 53, 70, 79, 87, 88, 99, 115, 117, 127, 128, 133, 147, 151, 163, 165, 172, 174, 177, 183, 185, 192, 193, 195, 197, 202, 211, 232, 242, 243, 244, 251, 258, 263, 265, 271, 276, 277, 278, 287, 289} \\ \midrule
      \rowcolor{cyan!10} \texttt{\textbf{Cluster 4 (Square Patterns in Fourier Magnitude)}} \\
    \rowcolor{cyan!10} \ttfamily{5, 7, 8, 12, 15, 18, 20, 22, 26, 27, 34, 37, 44, 45, 46, 52, 54, 55, 59, 63, 64, 66, 72, 73, 74, 76, 81, 89, 90, 92, 101, 103, 106, 108, 111, 116, 118, 122, 124, 129, 134, 135, 136, 139, 144, 145, 150, 152, 153, 156, 158, 160, 164, 168, 173, 176, 181, 182, 184, 188, 189, 190, 191, 194, 206, 208, 212, 213, 214, 219, 221, 222, 223, 224, 225, 228, 229, 230, 234, 236, 237, 238, 239, 248, 252, 253, 255, 259, 264, 270, 272, 273, 282, 284, 285, 286, 290, 293, 295, 299} \\ 
    \bottomrule
  \end{tabular}
  \vspace{-2em}
\end{table*}

\subsubsection{Image-to-Image Diffusion Models for Watermark Removal} We leveraged the \textbf{image-to-image} capabilities of the \textit{Stable Diffusion Refiner model}~\cite{meng2021sdedit} as a core component of our black-box watermark removal strategy. This approach exploits diffusion models' learned prior over natural images to project watermarked inputs onto the clean image manifold, effectively suppressing embedded signals while preserving semantic content. The process comprises two stages: (\textbf{i}) \emph{forward diffusion}, which progressively corrupts an image with Gaussian noise, and (\textbf{ii}) \emph{reverse diffusion}, where a denoising network reconstructs the image by iteratively removing noise. 

Given a watermarked image $x_w$, the forward process adds noise according to schedule $\{\alpha_t\}_{t=1}^T$, where strength parameter $s \in [0,1]$ controls the initial corruption level via $\alpha_1 = 1-s$ as
as $x_t = \sqrt{\alpha_t} x_w + \sqrt{1-\alpha_t} \epsilon_t$, where $\epsilon_t \sim \mathcal{N}(0, I)$. Here, the parameter $s$ critically balances watermark removal efficacy against content preservation: lower values ($s \approx 0.04$) maintain fine details but may insufficiently suppress watermarks, while higher values ($s \approx 0.25$) remove watermarks more aggressively at the risk of semantic drift (see Figure~\ref{fig:img2img}).
Subsequently, the \textit{reverse diffusion process} iteratively removes noise through a learned denoising function $\epsilon_\theta$, reconstructing the final refined image as 
$x_{t-1} = \frac{1}{\sqrt{\alpha_t}} \left( x_t - \sqrt{1-\alpha_t} \epsilon_\theta(x_t, t) \right)$.

We configured the diffusion process with $T = 500$ inference steps to enable sufficient iterative refinement for thorough watermark suppression. To preserve semantic content, we further introduced \textbf{semantic guidance} via ChatGPT-4-generated captions~\cite{achiam2023gpt} that provide accurate, content-specific descriptions of each image (\textit{e.g.}, \textit{``a medieval castle on a hilltop surrounded by forests''}). These captions anchor the denoising trajectory, ensuring that reconstructed regions align with natural image statistics rather than residual watermark patterns. The classifier-free guidance scale was set to $w = 1.0$ to maintain photorealism: higher values ($w > 2$) increase text condition strength but tend to introduce stylization artifacts inconsistent with the original content~\cite{song2025rethinking}. By tuning the strength parameter $s$ for each cluster, we achieved a balanced trade-off between aggressive watermark removal and high perceptual fidelity. 
\begin{wrapfigure}{r}{0.32\textwidth} 
  \centering
  \vspace{2em}
  \includegraphics[clip, trim=10 10 10 10, width=0.35\textwidth]{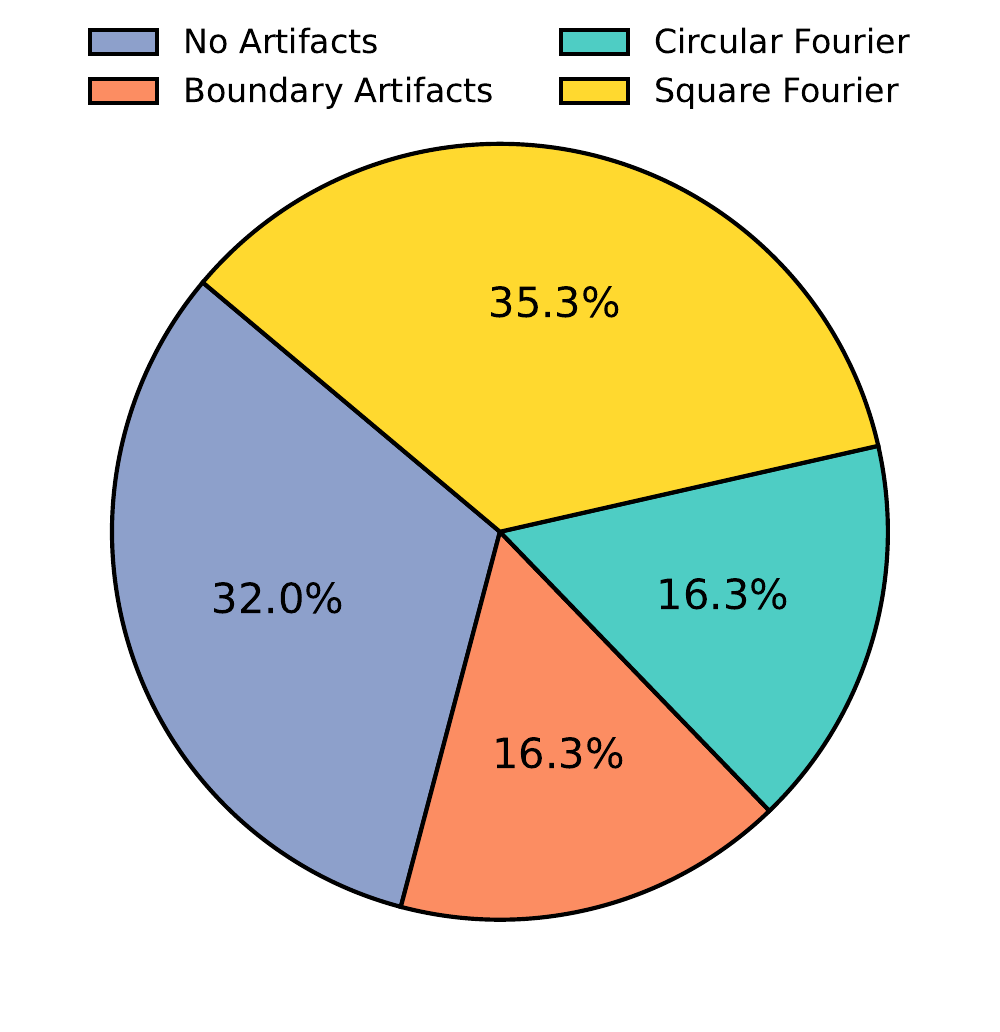}
  \caption{\small Images distribution for Black-box track clusters.}
  \label{fig:distribution}
  \vspace{-3em}
\end{wrapfigure}

\subsection{Cluster-Specific Solutions} \label{sec:cluster-specific}
Our empirical clustering analysis revealed distinct artifact patterns, enabling the design of \textbf{adaptive, cluster-tailored watermark removal strategies} rather than relying on a single, global configuration. 
This specialization significantly improved both perceptual quality and watermark suppression efficacy.  

\noindent {\color{teal}\textbf{Cluster 1: No discernible artifacts}.}  
In the absence of visible spatial or frequency-domain cues, we adopted a high-strength \textit{image-to-image diffusion} configuration ($s = 0.16$). This choice prioritizes aggressive denoising to suppress potential low-energy watermark embeddings, while relying on semantic guidance to recover fine details without introducing perceptual distortions.

\noindent {\color{olive}\textbf{Cluster 2: Boundary artifacts}.}  
For images exhibiting border-like spatial patterns, we employed a \textbf{three-stage removal pipeline}, as in the Beige-box track (detailed in Sec.~\ref{sec:stegastamp}):  
\begin{enumerate}[label=(\roman*)]
    \item \textit{Paired Synthetic dataset generation}, where each pair consists of two images watermarked with the same spatial pattern but opposite bit sequences.  
    \item \textit{VAE fine-tuning} to minimize reconstruction loss between watermarked inputs and clean target images, thereby learning an implicit watermark removal prior.  
    \item \textit{Post-processing enhancement} via CIELAB-space color transfer and local contrast adjustment to restore natural appearance without reintroducing watermark signals.  
\end{enumerate}  

\noindent {\color{orange}\textbf{Cluster 3: Circular patterns in the Fourier spectrum}.} This cluster was treated using the same three-stage beige-box track pipeline as {\color{olive}\textbf{Cluster 2}}.

\noindent {\color{blue}\textbf{Cluster 4: Square patterns in the Fourier spectrum}.}  
We adopted a \textbf{hybrid approach} combining low-strength ($s = 0.04$) \textit{image-to-image diffusion} to minimally perturb fine details, followed by a 7-pixel horizontal translation with selective restoration, similar to the TreeRing watermark removal method (detailed in Sec.~\ref{sec:treering}).  

\noindent Across clusters, hyperparameters (especially $s$) were tuned to balance \textbf{removal aggressiveness vs. content fidelity}. This cluster-specific methodology outperformed uniform parameter settings, yielding notable improvements in both qualitative assessments and quantitative metrics. Details for Black-box approach are provided in Algorithm~\ref{alg:blackbox-pipeline}.


\begin{algorithm}[t]
\caption{{\color{purple}\textbf{Black–Box Track Pipeline}}}
\label{alg:blackbox-pipeline}
\begin{algorithmic}[1]
  \Require Watermarked images $\{x_w\}$, manual cluster labels $\{c_x\in\{1,2,3,4\}\}$
  \vspace{1ex}
  \State \textcolor{StageColor}{\texttt{\#Stage 1: Manual Clustering}}
  \State \quad Assign each image $x\in\{x_w\}$ to cluster $c_x$ by visual inspection:
  \State \quad \quad \texttt{\textbf{Cluster 1}}: Boundary artifacts \quad\quad\quad  \texttt{\textbf{Cluster 2}}: Circular Fourier patterns 
  \State \quad \quad \texttt{\textbf{Cluster 3}}: Square Fourier patterns  \quad \texttt{\textbf{Cluster 4}}: No noticeable artifacts

  \vspace{1ex}
  \State \textcolor{StageColor}{\texttt{\#Stage 2: Cluster‐Specific Solutions}}
  \ForAll{$x$ \textbf{in} $\{x_w\}$}
    \State $k \gets c_x$
    \If{$k = 1$}
      \State \textcolor{KeywordColor}{\# Diffusion: aggressive removal}
      \State $x_r \gets \mathrm{Diffuse}(x,\;s=0.16; \text{ChatGPT captions})$\Comment{Image-to-image diffusion pipeline}
    \ElsIf{$k \in \{2,3\}$}
      \State \textcolor{KeywordColor}{\# VAE Pipeline (Sec.~\ref{sec:stegastamp})}
      \State $x_r \gets \mathrm{VAE\_remove}(x)$\Comment{Includes dataset generation, finetuning, color–contrast}
    \ElsIf{$k = 4$}
      \State \textcolor{KeywordColor}{\# Hybrid: mild diffusion + translation}
      \State $x_d \gets \mathrm{Diffuse}(x,\;s=0.04)$
      \State $x_s \gets \mathrm{Translate}(x_d,\;7)$\Comment{Image translation by 7 pixels}
      \State Restore leftmost 7 columns from $x$ into $x_s$
      \State $x_r \gets x_s$
    \EndIf
    \State save $x_r$  \Comment{Refined output}
  \EndFor

  \Ensure Refined images $\{x_r\}$
\end{algorithmic}
\end{algorithm}
\begin{figure*}[t]
    \centering
    \vspace{-2pt} 
    \includegraphics[clip, trim=5 5 5 5, width=\textwidth]{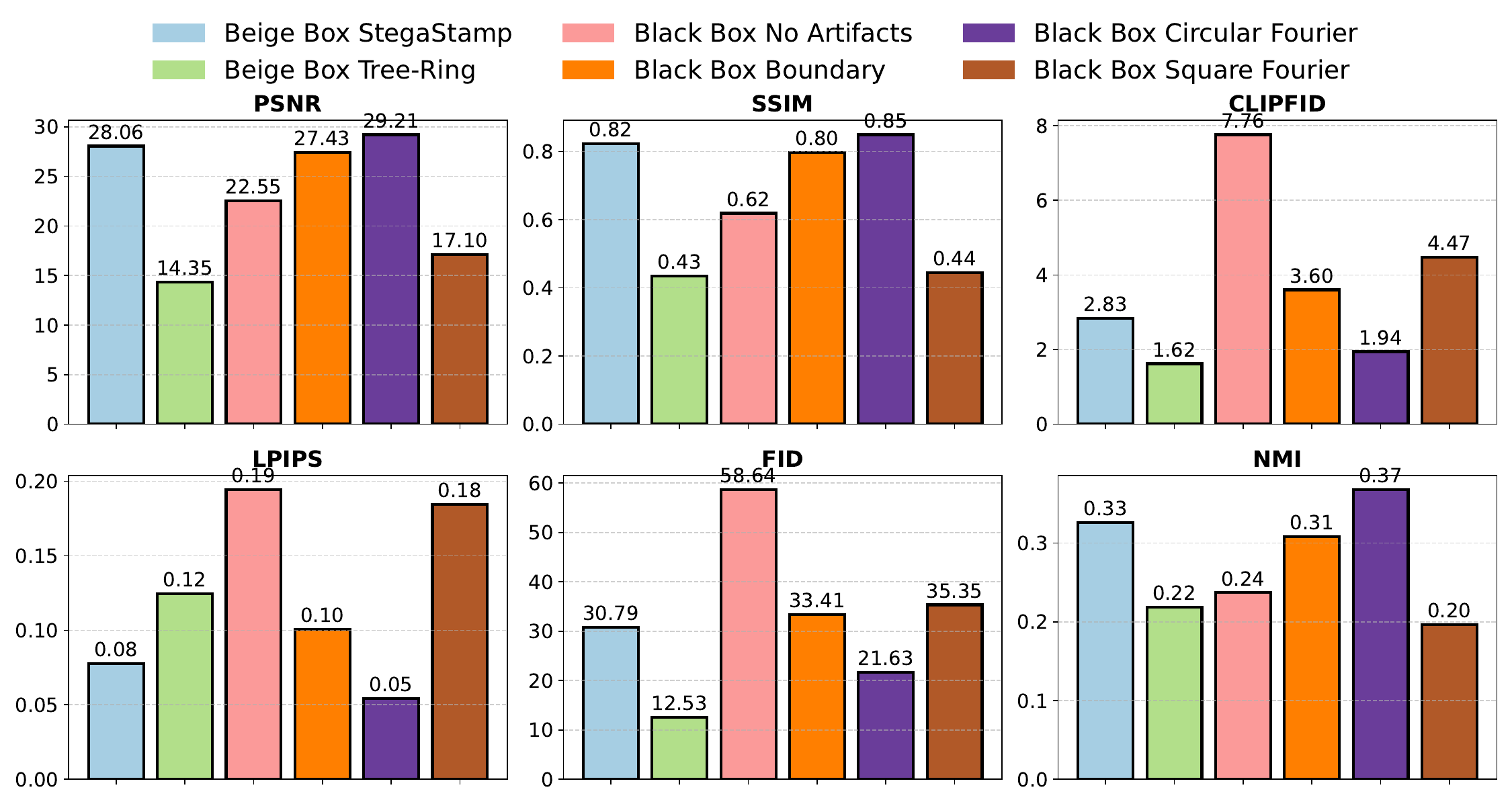}
    \caption{\small \textbf{Performance of our pipeline across beige-box and black-box tracks}. StegaStamp achieves the most balanced quality–removal trade-off, aided by CIELAB-based post-processing, while TreeRing and the black-box cluster with Fourier square patterns show reduced score for alignment metrics (PSNR, SSIM) due to the intentional translation shift.}
    \label{fig:metric_comp}
    \vspace{-7pt} 
\end{figure*}

\begin{figure*}[t]
    \centering
    \vspace{-2pt} 
    \includegraphics[clip, trim=5 5 5 5, width=\textwidth]{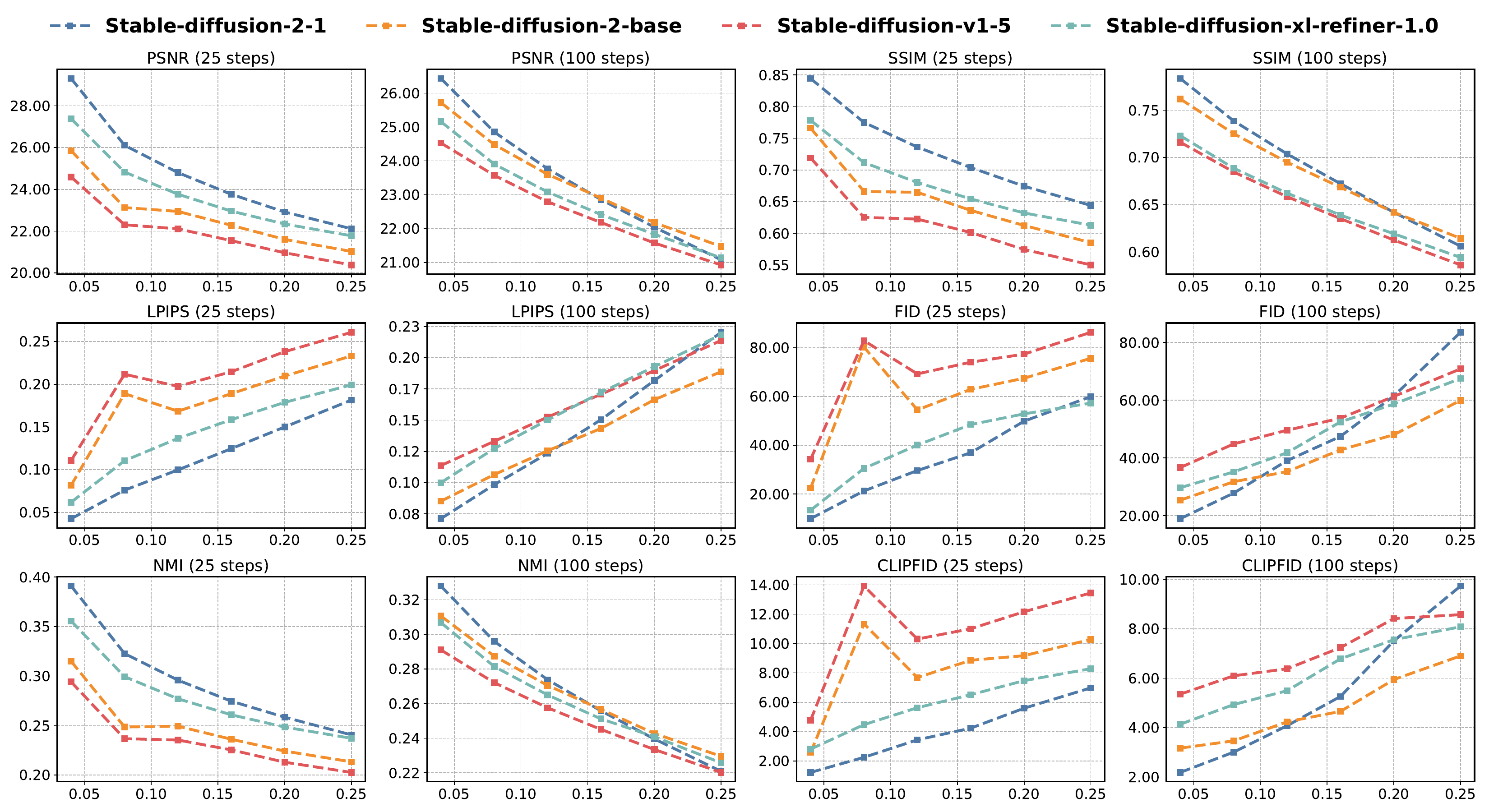}
    \caption{\small  \textbf{Effect of strength $s$ and steps $T$ on watermark removal}. PSNR drops sharply at $s \approx 0.10$, motivating cluster-specific settings. From leaderboard results, we observe that \texttt{stable-diffusion-xl-refiner-1.0} consistently outperforms other diffusion models in terms of watermark removal.}
    \label{fig:diffusion-models}
    \vspace{-7pt} 
\end{figure*}

\section{Additional Experiments}  

To further understand the trade-off between watermark removal strength and perceptual quality, we conducted a set of controlled experiments across different diffusion models and sampling configurations. These results, while not part of the main leaderboard evaluations, provide quantitative support for the hyperparameter choices.  

\textbf{Quality Metrics Trends Across Tracks and Clusters.} Figure~\ref{fig:metric_comp} evaluates our optimized pipeline across both beige-box tracks and four black-box clusters. The results highlight distinct quality--removal trade-offs. \textit{Beige-box StegaStamp} achieves the most balanced outcome (PSNR: 28.06$\pm$1.2, SSIM: 0.82$\pm$0.04), with high semantic fidelity (CLIP-FID $<$ 5) and strong pixel-wise alignment---attributable in part to our CIELAB-based post-processing stage, which restores fine-scale color consistency after watermark removal. In contrast, \textit{TreeRing} and \textit{Black-box Cluster with square Fourier artifacts} show markedly lower PSNR and SSIM, a direct consequence of the 7-pixel horizontal translation that disrupts watermark embeddings; despite this, semantic similarity remains high (CLIP-FID $<$ 5). Finally, \textit{Black-box Cluster with no artifacts} exhibits the largest perceptual gap, indicating that its imperceptible watermark necessitated more aggressive interventions, with some loss of fine-grained detail.

\textbf{Impact of Diffusion Strength and Model Choice.}

Figure~\ref{fig:diffusion-models} presents evaluation of four Stable Diffusion variants across strength parameters $s \in [0.04, 0.25]$ and inference steps $T \in \{25, 100\}$. Quality degradation exhibits a sharp nonlinearity at $s \approx 0.10$, where PSNR drops from 26 to 22\,dB, motivating our conservative $s=0.04$ for structured artifacts versus aggressive $s=0.16$ for hidden watermarks. Similarly, extended denoising (100 vs.\ 25 steps) yields marginal improvements ($\sim$ 0.5\,dB PSNR, 0.02 LPIPS). In addition, from our leaderboard submissions we observe that \texttt{stable-diffusion-xl-refiner-1.0} is more effective at watermark removal than other tested diffusion models.

\section{Exploratory but Ineffective Approaches}
\begin{figure*}[t]
   \centering
   \includegraphics[clip, trim=0 0 0 0, width=0.95\textwidth]{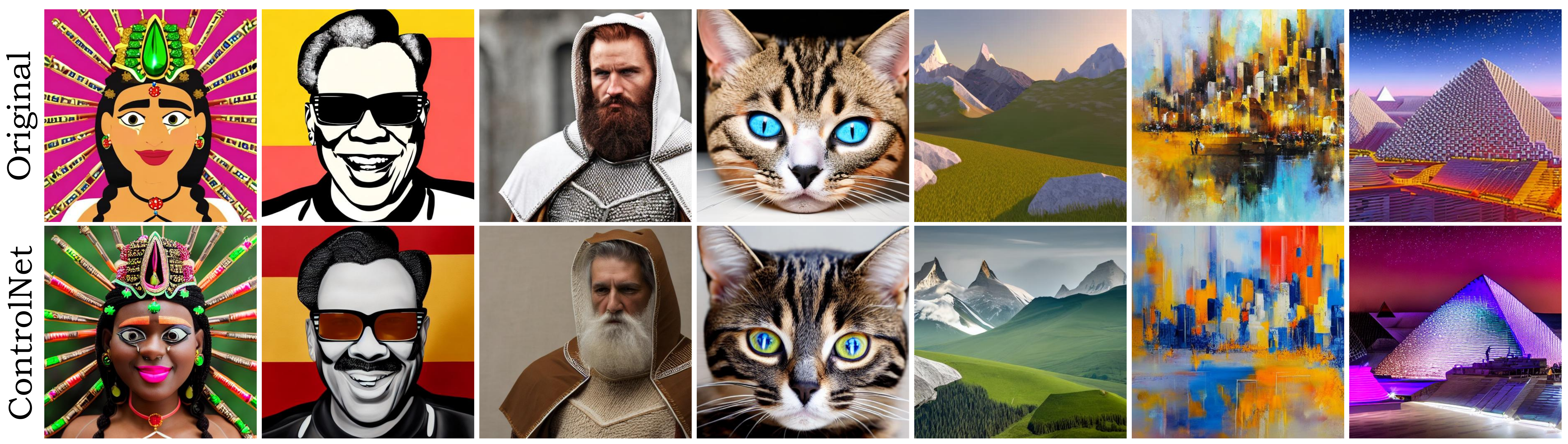} 
   \vspace{-0em}
   \caption{Qualitative results of the ControlNet-based edge guidance watermark removal approach. \textit{Top row}: original watermarked images. \textit{Bottom row}: ControlNet-generated outputs using Canny edges and color-aware captions.}
   \vspace{-1em}
   \label{fig:controlnet} 
\end{figure*}
\begin{wrapfigure}{r}{0.6\textwidth} 
  \centering
  \includegraphics[clip, trim=5 5 5 5, width=0.55\textwidth]{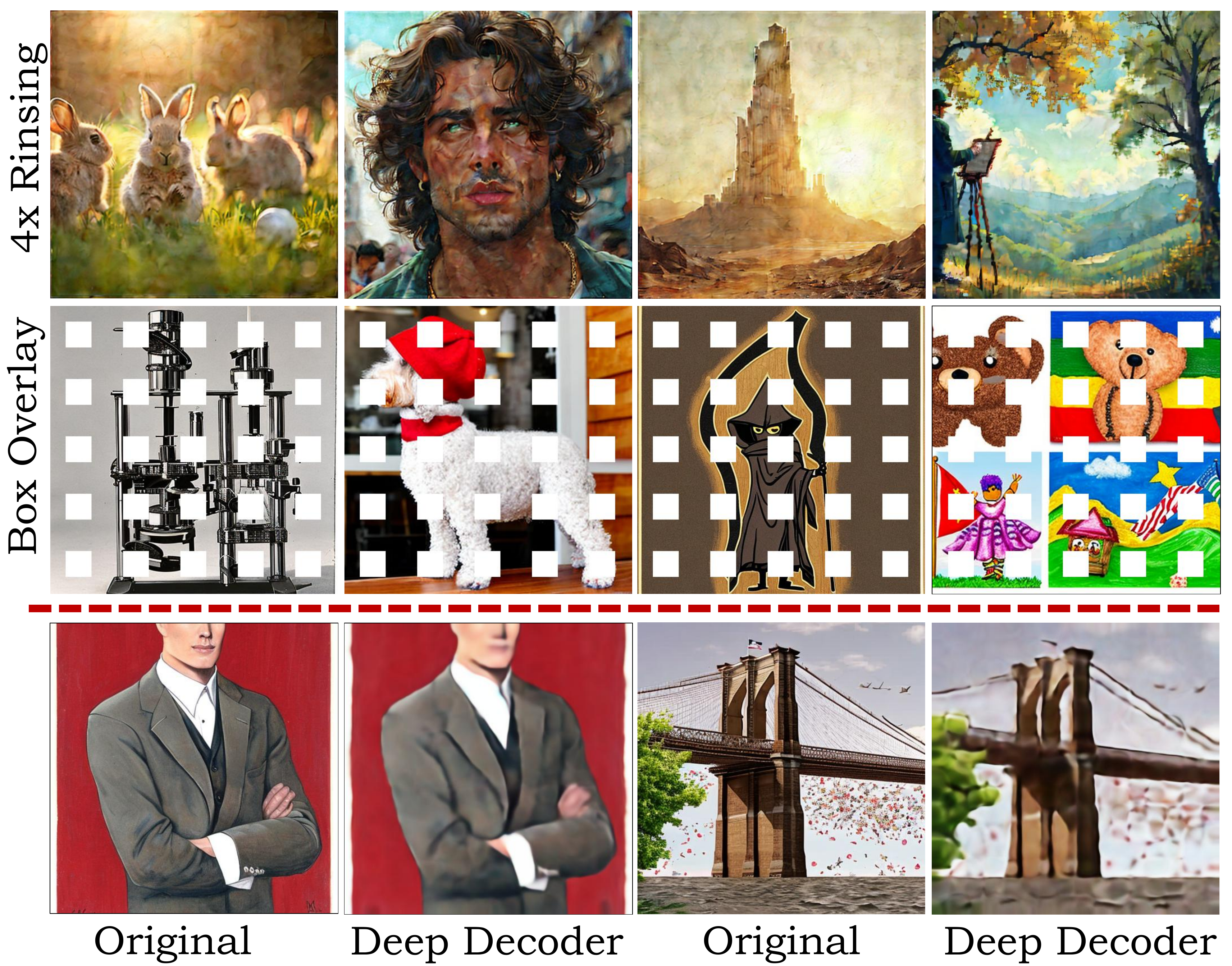}
  \caption{Qualitative examples of failed watermark removal strategies. \textbf{\textit{Row 1}}: aggressive four-pass latent diffusion rinsing is effective at watermark suppression but severely degrading perceptual quality. \textbf{\textit{Row 2}}: white-box additive perturbations results in limited removal and noticeable texture artifacts. \textbf{\textit{Row 3}}: untrained neural network reconstruction removes most watermarks but introduces oversmoothing and color shifts. Best viewed zoomed in.}
  \vspace{-1em}
  \label{fig:failed-app}
\end{wrapfigure}
In parallel to developing our primary solution, we systematically explored multiple alternative strategies for watermark removal. While each was grounded in plausible hypotheses, none yielded competitive results in terms of the removal–quality trade-off. Below, we summarize these unsuccessful attempts along with representative qualitative outcomes.

\textbf{Watermark Removal via ControlNet-Based Edge Guidance.} We explored a ControlNet-based~\cite{zhang2023adding} strategy where the Canny edges of each watermarked image were extracted and used as structural guidance. To provide semantic conditioning, we generated descriptive captions using ChatGPT, explicitly incorporating both scene content and color attributes (\textit{e.g.}, \textit{“a red vintage car parked in front of a rustic wooden cabin, surrounded by snow”}). These captions aimed to preserve fine-grained semantic and chromatic fidelity in the regenerated output. The edge maps and captions were then fed into ControlNet with Stable Diffusion to synthesize new images that aligned closely with the original structure and color scheme.
Despite producing visually coherent outputs, as shown in Figure~\ref{fig:controlnet}, the watermarks were largely retained. This outcome aligns with recent findings that watermark signal components often survive structural-guided regeneration when the conditioning enforces strong pixel-level consistency with the original image~\cite{lukovnikovsemantic}. In the qualitative example, the top row shows the original watermarked images, and the bottom row shows the ControlNet-generated outputs, which maintain the global scene layout but fail to remove the embedded watermark.

\textbf{Aggressive Diffusion Rinsing.}
We applied aggressive “diffusion rinsing,”,where the image is repeatedly passed through forward–reverse diffusion cycles in latent space, effectively purifying the image by destroying and reconstructing its content multiple times. A four-pass rinsing cycle removed most visible watermark traces; however, as shown in the first row of Figure \ref{fig:failed-app}, this came at the cost of severe perceptual degradation, with washed-out textures and loss of fine details. Pixel-space diffusion models were also tested but did not yield noticeable improvements in removal performance.

\textbf{White-Boxes Overlay.}
In the early competition phase, we experimented with directly adding white-box perturbations to the input images with the aim of disrupting watermark patterns. This not only failed to erase the  watermarks but also caused quality degradation, producing visually unnatural textures, as evident in the second row of Figure \ref{fig:failed-app}.

\textbf{Untrained Neural Network Reconstruction.}
Motivated by the premise that untrained neural networks capture natural image priors~\cite{ulyanov2018deep}, we attempted to reconstruct the original watermarked images from scratch using an untrained convolutional decoder. This approach succeeded in attenuating most watermarks, but reconstruction artifacts, oversmoothed textures, and loss of color fidelity were evident, as seen in the third row of Figure \ref{fig:failed-app}. The method’s inherent slow convergence and inability to perfectly reproduce fine structures limited its practicality for competition-scale deployment.

\textbf{Additional Unsuccessful Attempts.}
Beyond the above failed strategies, we explored several other methods that ultimately proved ineffective. We implemented \textit{adversarial attacks on an ensemble of latent-space encoders}, following the methodology outlined in \texttt{ding2024waves}, but these did not disrupt the embedded watermarks in a meaningful way. We also attempted \textit{super-resolution and inpainting-based perturbations}, hypothesizing that re-synthesizing high-frequency details or contextually filling regions might break watermark structures; however, these approaches only introduced visual artifacts without consistent removal. Other tested methods included \textit{color space perturbations}, \textit{frequency domain masking}, and \textit{hybrid denoising–sharpening pipelines}, all of which degraded perceptual quality while leaving partially intact watermark traces.

\section{Conclusion}

In conclusion, our work provides a comprehensive examination of the vulnerabilities in current invisible watermarking schemes, demonstrating that even under restrictive threat models, including the challenging black-box setting, determined attackers can effectively strip watermarks while preserving perceptual quality. Our results, which topped \textit{Beige-box and Black-box NeurIPS 2024 competition tracks}, highlight not only the feasibility but also the generality of such removal strategies across fundamentally different watermark methods. This calls for a rethinking of watermark robustness, emphasizing the urgent need for schemes that can withstand adaptive, high-capacity generative manipulation without sacrificing utility, thereby ensuring reliable provenance tracking in the era of powerful image synthesis models.

\section*{Acknowledgment}

We would like to thank Mucong Ding, Tahseen Rabbani, Bang An, and Chenghao Deng for their consistent availability and support throughout the competition, particularly in resolving Codabench-related issues. We are also grateful to Tom Goldstein and Furong Huang for their efforts in organizing the NeurIPS 2024 watermarking competition and for fostering a platform to advance research on robustness and provenance in generative models.

\bibliography{iclr2025}
\bibliographystyle{iclr2025_conference}

\clearpage
\input{supplementary}

\end{document}

%% file: supplementary.tex
\clearpage

\section*{Appendix} 

This appendix provides background information and extensive qualitative results.  
For clarity, we summarize the contents and their page numbers below.

\vspace{1em}

\newcommand{\AppLine}[3]{%
  \noindent\textbf{#1}\quad #2%
  \dotfill\makebox[2em][r]{~\pageref{#3}}\par
}

\begin{figure*}[t]
    \centering
    \vspace{-2em} 
    \includegraphics[clip, trim=0 0 0 0, width=\textwidth]{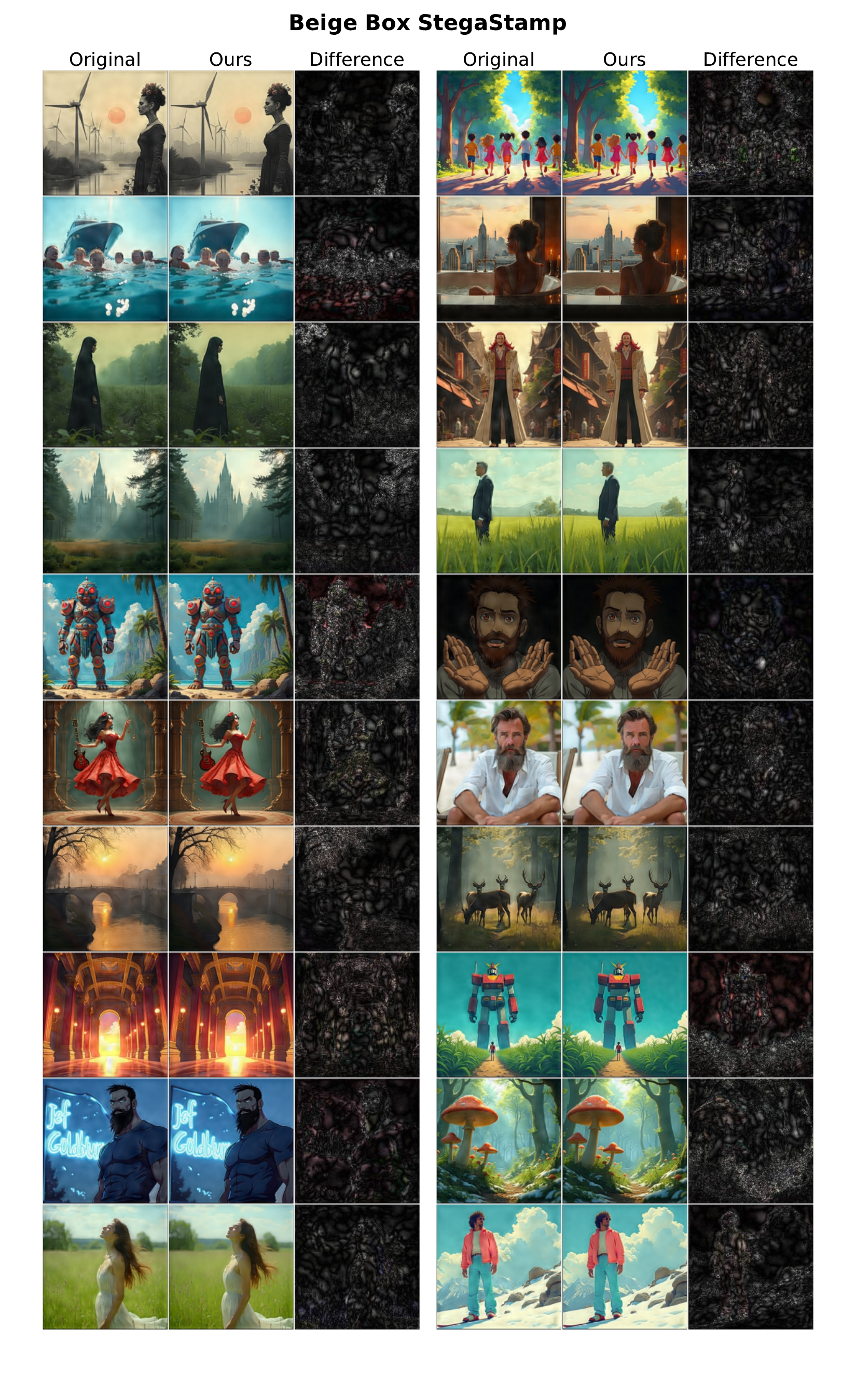}
    \vspace{-5em} 
    \caption{\small  Qualitative results for watermark removal on StegaStamp images in the Beige-box track. Each triplet shows (left) the original watermarked input, (middle) the output after applying our removal method, and (right) the residual difference between them. Our approach preserves semantic content and visual fidelity while effectively eliminating the embedded watermark.}
    \label{fig:diff_stega}
\end{figure*}

\begin{figure*}[t]
    \centering
    \vspace{-2em} 
    \includegraphics[clip, trim=0 0 0 0, width=\textwidth]{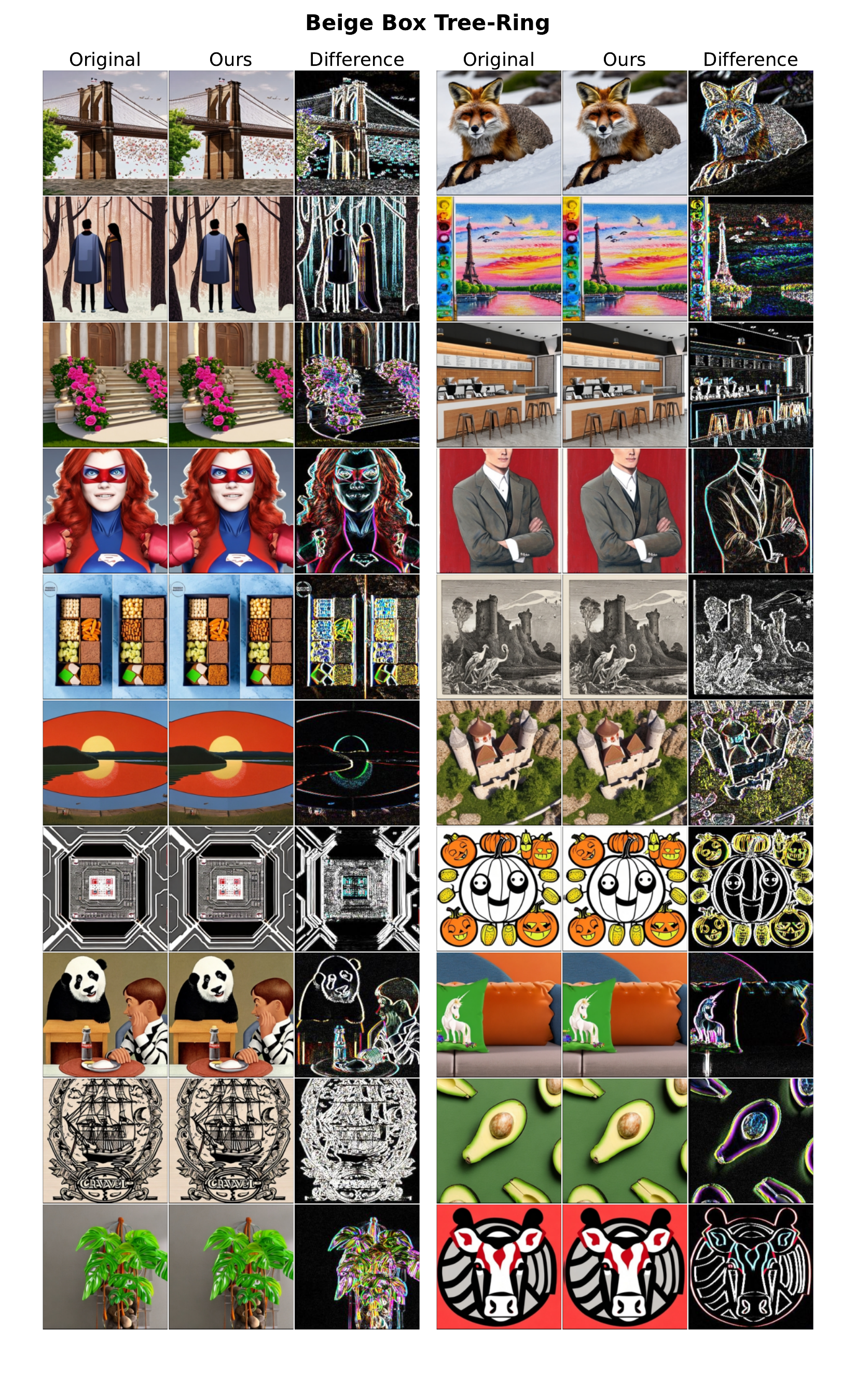}
    \vspace{-5em} 
    \caption{\small  Qualitative results for watermark removal on TreeRings images in the Beige-box track. Each triplet shows (left) the original watermarked input, (middle) the output after applying our removal method, and (right) the residual difference between them. Our approach preserves semantic content and visual fidelity while effectively eliminating the embedded watermark.}
    \label{fig:diff_treering}
\end{figure*}

\begin{figure*}[t]
    \centering
    \vspace{-2em} 
    \includegraphics[clip, trim=0 0 0 0, width=\textwidth]{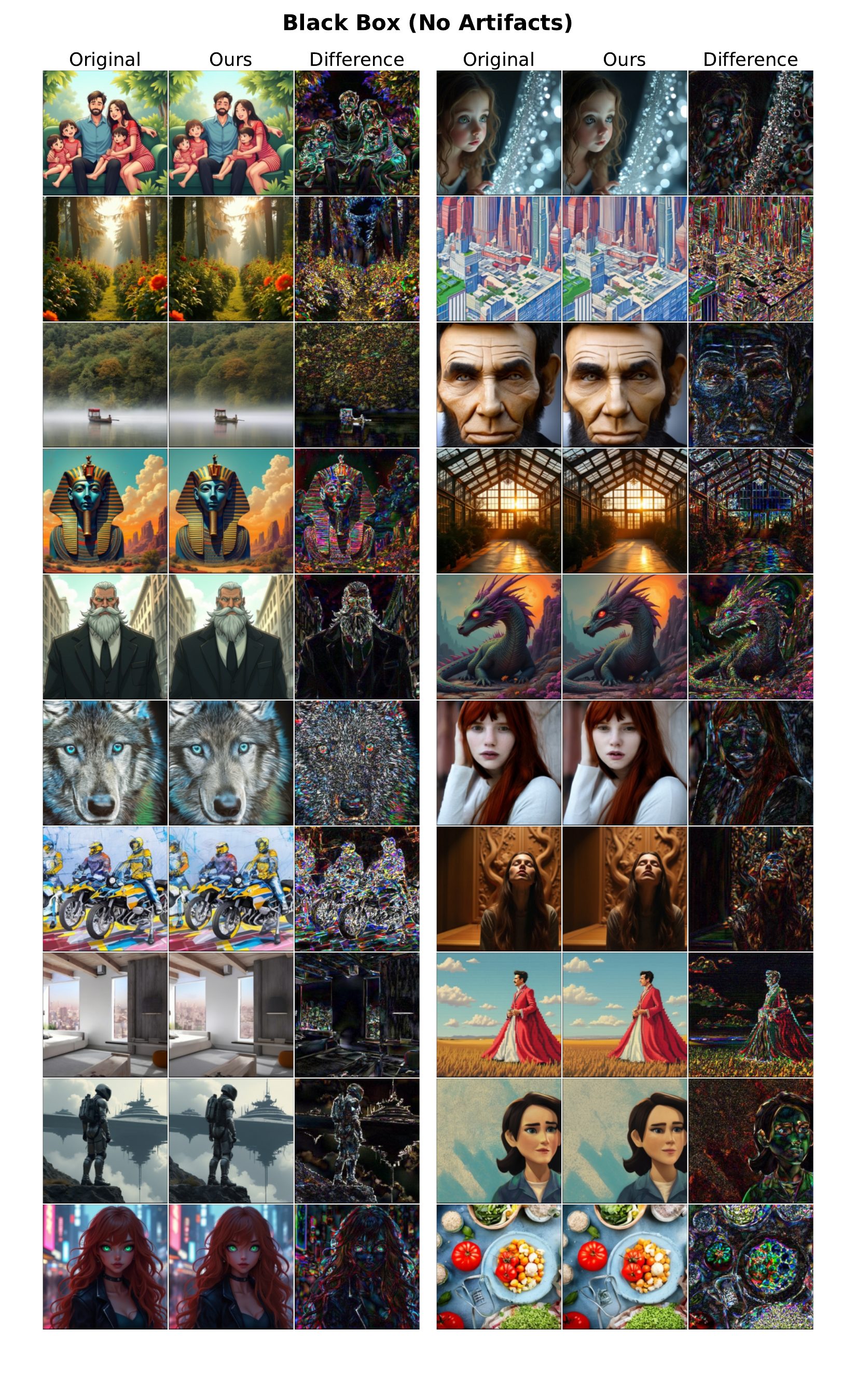}
    \vspace{-5em} 
    \caption{\small  Qualitative results on the Black-box track for images without noticeable artifacts. Each triplet shows (left) the original watermarked input, (middle) the output after applying our removal method, and (right) the residual difference between them. Our approach preserves semantic content and visual fidelity while effectively eliminating the embedded watermark.}
    \label{fig:diff_bbnoartifact}
\end{figure*}

\begin{figure*}[t]
    \centering
    \vspace{-2em} 
    \includegraphics[clip, trim=0 0 0 0, width=\textwidth]{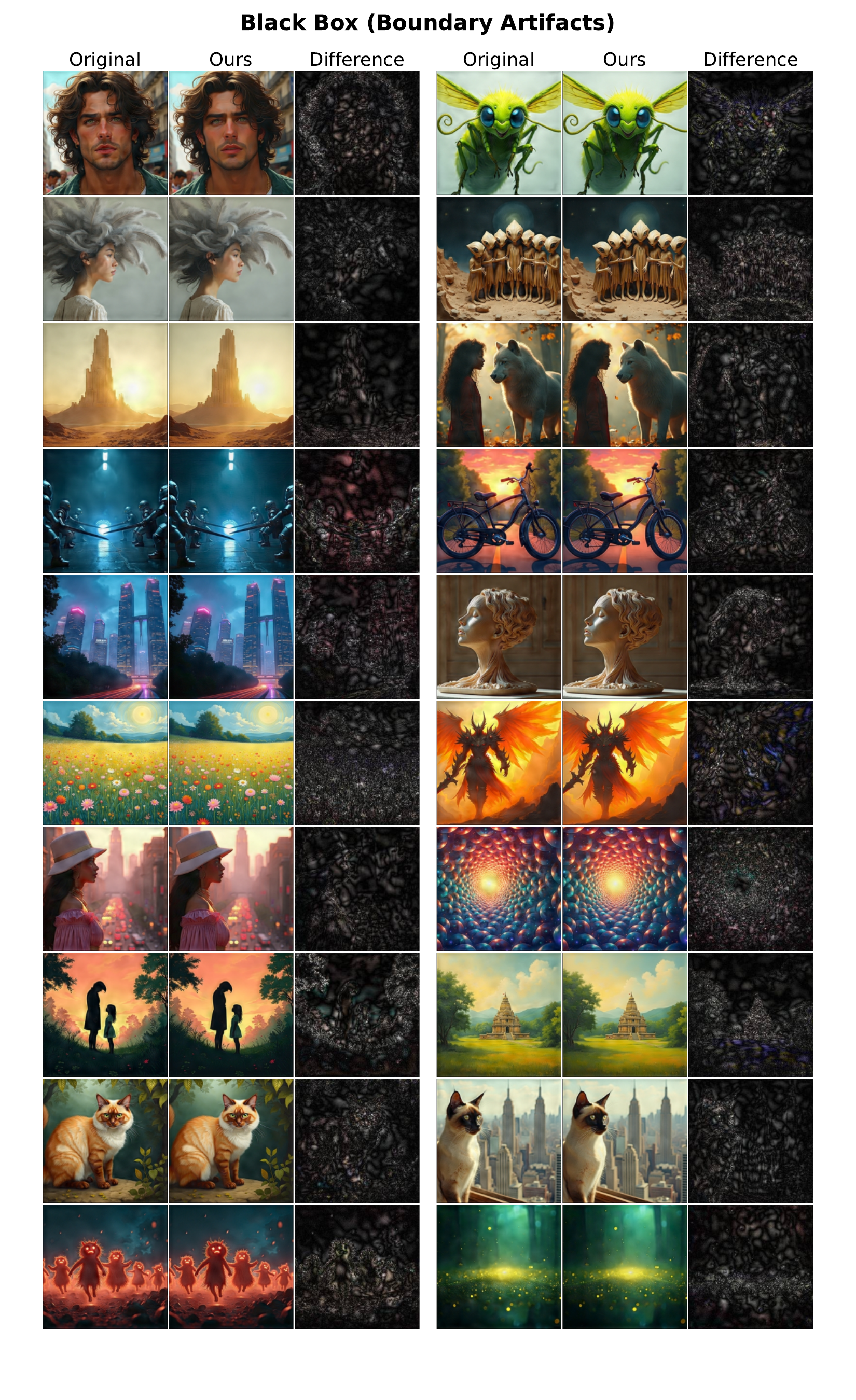}
    \vspace{-5em} 
    \caption{\small  Qualitative results on the Black-box track for images with boundary artifacts. Each triplet shows (left) the original watermarked input, (middle) the output after applying our removal method, and (right) the residual difference between them. Our approach preserves semantic content and visual fidelity while effectively eliminating the embedded watermark.}
    \label{fig:diff_bbboundary}
\end{figure*}

\begin{figure*}[t]
    \centering
    \vspace{-2em} 
    \includegraphics[clip, trim=0 0 0 0, width=\textwidth]{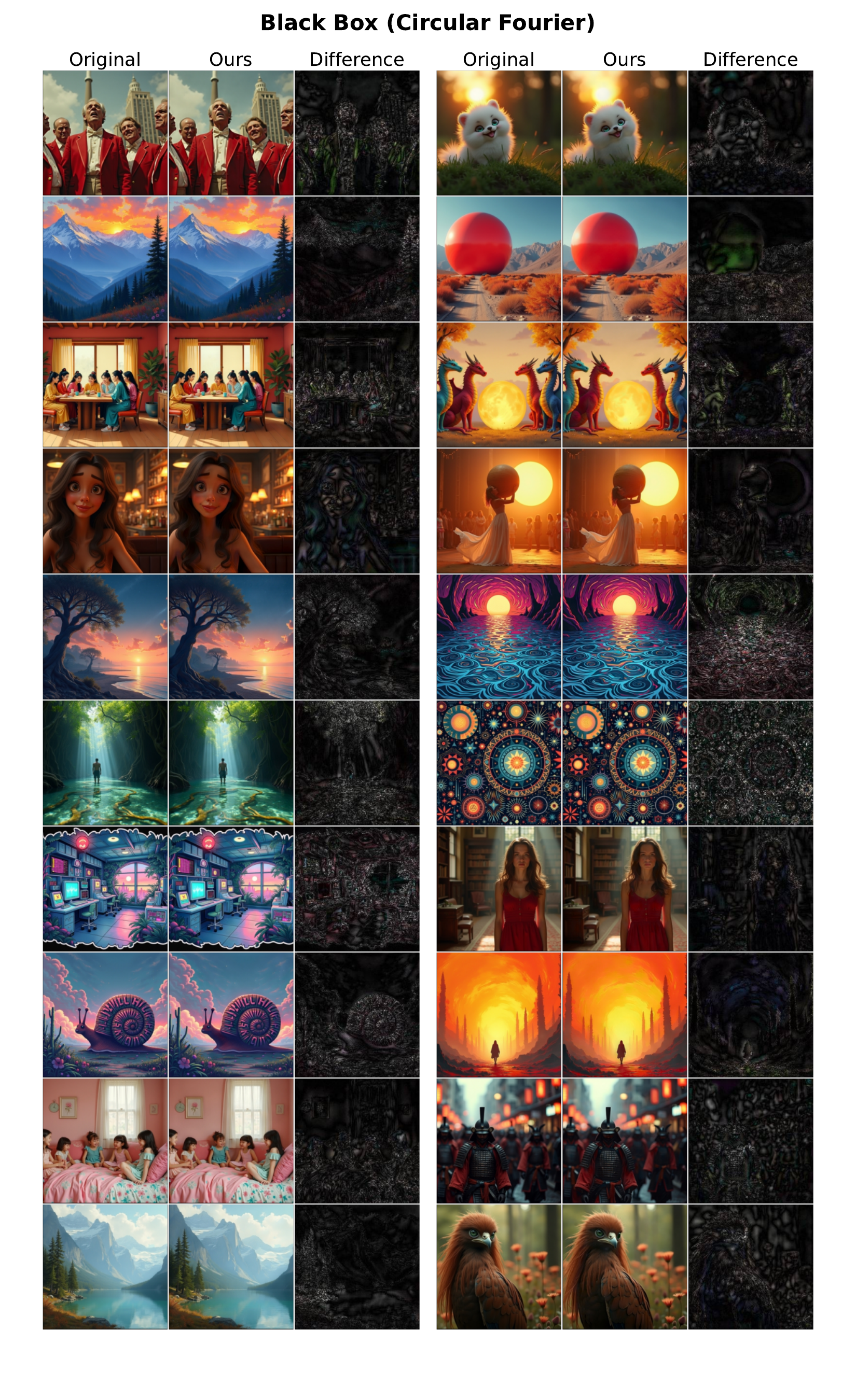}
    \vspace{-5em} 
    \caption{\small  Qualitative results on the Black-box track for images containing circular Fourier-domain artifacts. Each triplet shows (left) the original watermarked input, (middle) the output after applying our removal method, and (right) the residual difference between them. Our approach preserves semantic content and visual fidelity while effectively eliminating the embedded watermark.}
    \label{fig:diff_bbcirc}
\end{figure*}

\begin{figure*}[t]
    \centering
    \vspace{-2em} 
    \includegraphics[clip, trim=0 0 0 0, width=\textwidth]{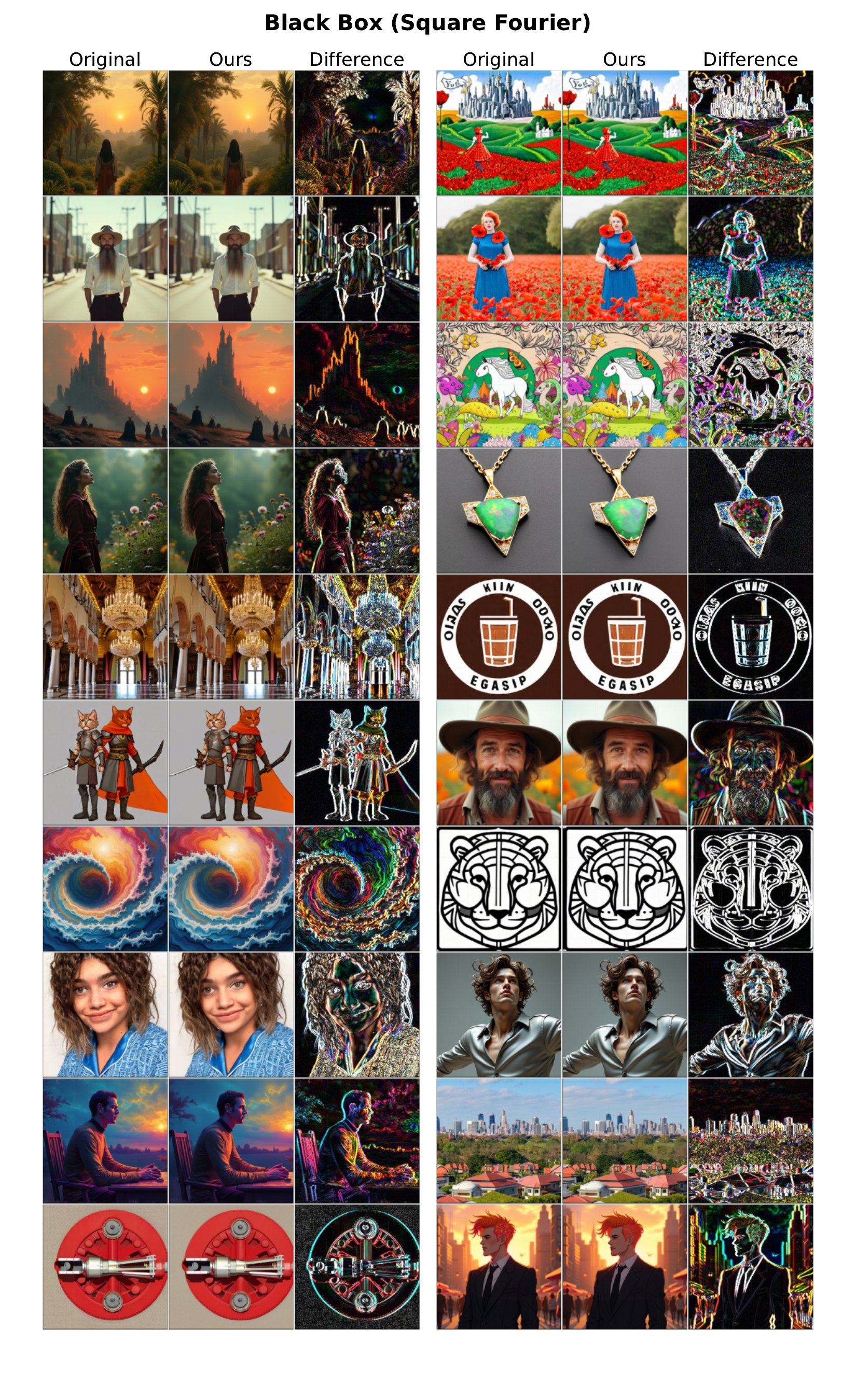}
    \vspace{-5em} 
    \caption{\small  Qualitative results on the Black-box track for images containing square Fourier-domain artifacts. Each triplet shows (left) the original watermarked input, (middle) the output after applying our removal method, and (right) the residual difference between them. Our approach preserves semantic content and visual fidelity while effectively eliminating the embedded watermark}
    \label{fig:diff_bbsquare}
\end{figure*}

\begin{figure*}[t]
    \centering
    \vspace{-2em} 
    \includegraphics[clip, trim=0 0 0 0, width=\textwidth]{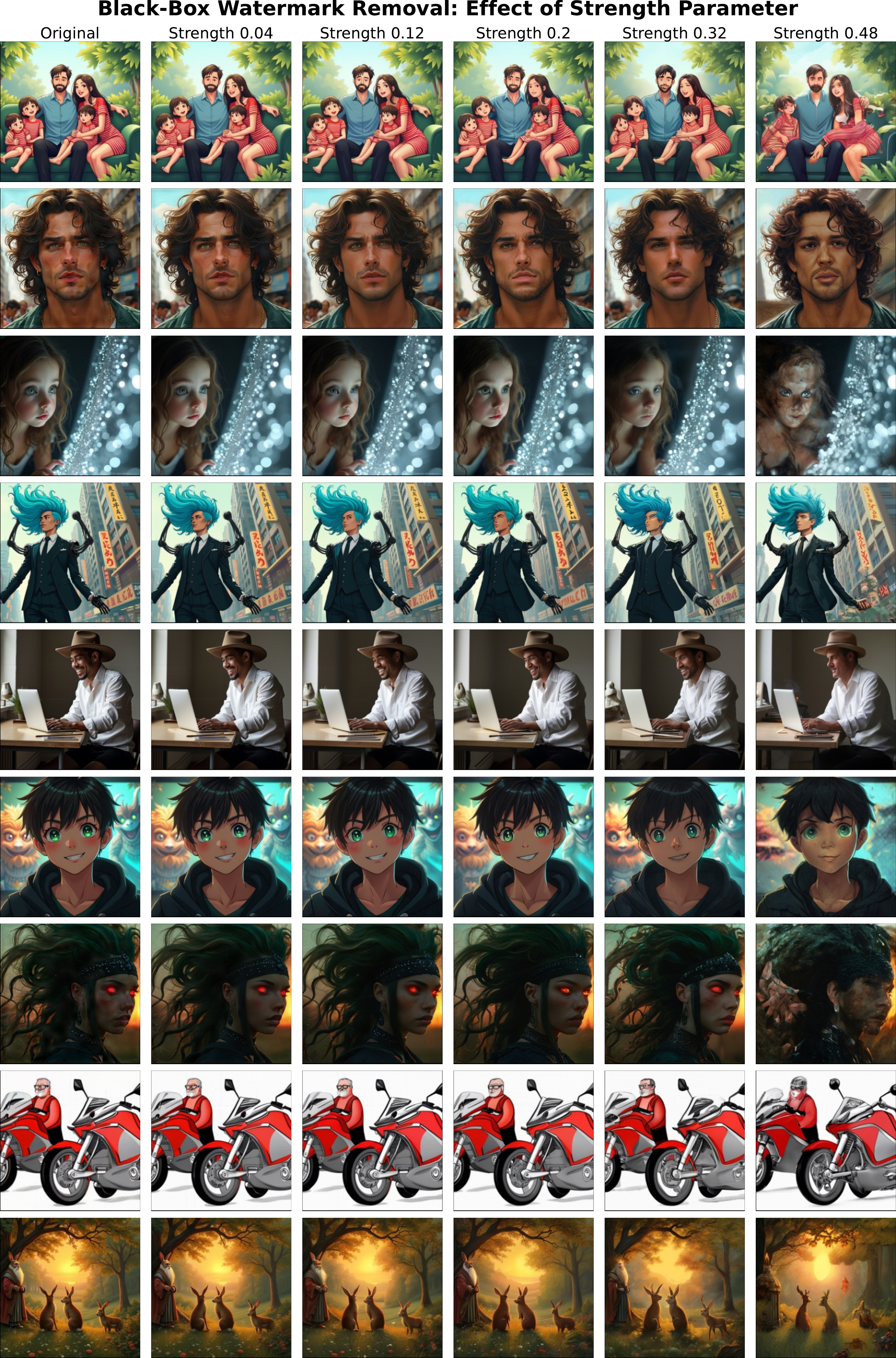}
    \vspace{-1em} 
    \caption{\small  Effect of varying the diffusion strength parameter on Black-box watermark removal. Lower values (0.04–0.12) maintain high perceptual quality with partial removal, whereas higher values yield stronger removal at the cost of visible distortions. This illustrates the quality–robustness trade-off inherent in diffusion-based purification.}
    \label{fig:diff_bbstrength}
\end{figure*}

\begin{figure*}[t]
    \centering
    \vspace{-2em} 
    \includegraphics[clip, trim=65 45 10 25, width=\textwidth]{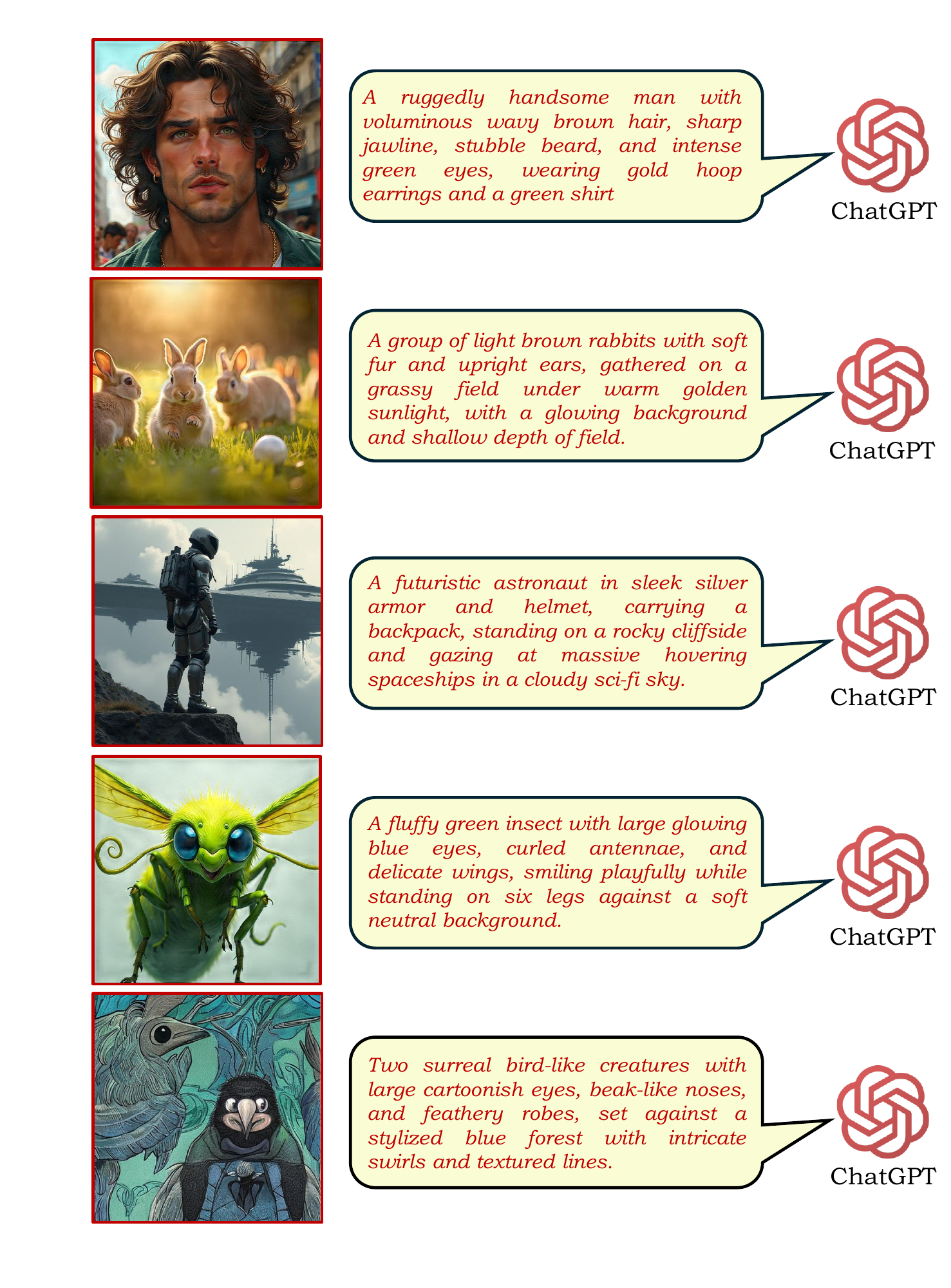}
    \vspace{-1em} 
    \caption{\small  Captions automatically generated by ChatGPT for a set of original watermarked images. These textual descriptions capture semantic and stylistic details of the inputs and are subsequently used to guide image-to-image diffusion for watermark removal.}
    \label{fig:diff_bbchat1}
\end{figure*}

\begin{figure*}[t]
    \centering
    \vspace{-2em} 
    \includegraphics[clip, trim=65 45 10 25, width=\textwidth]{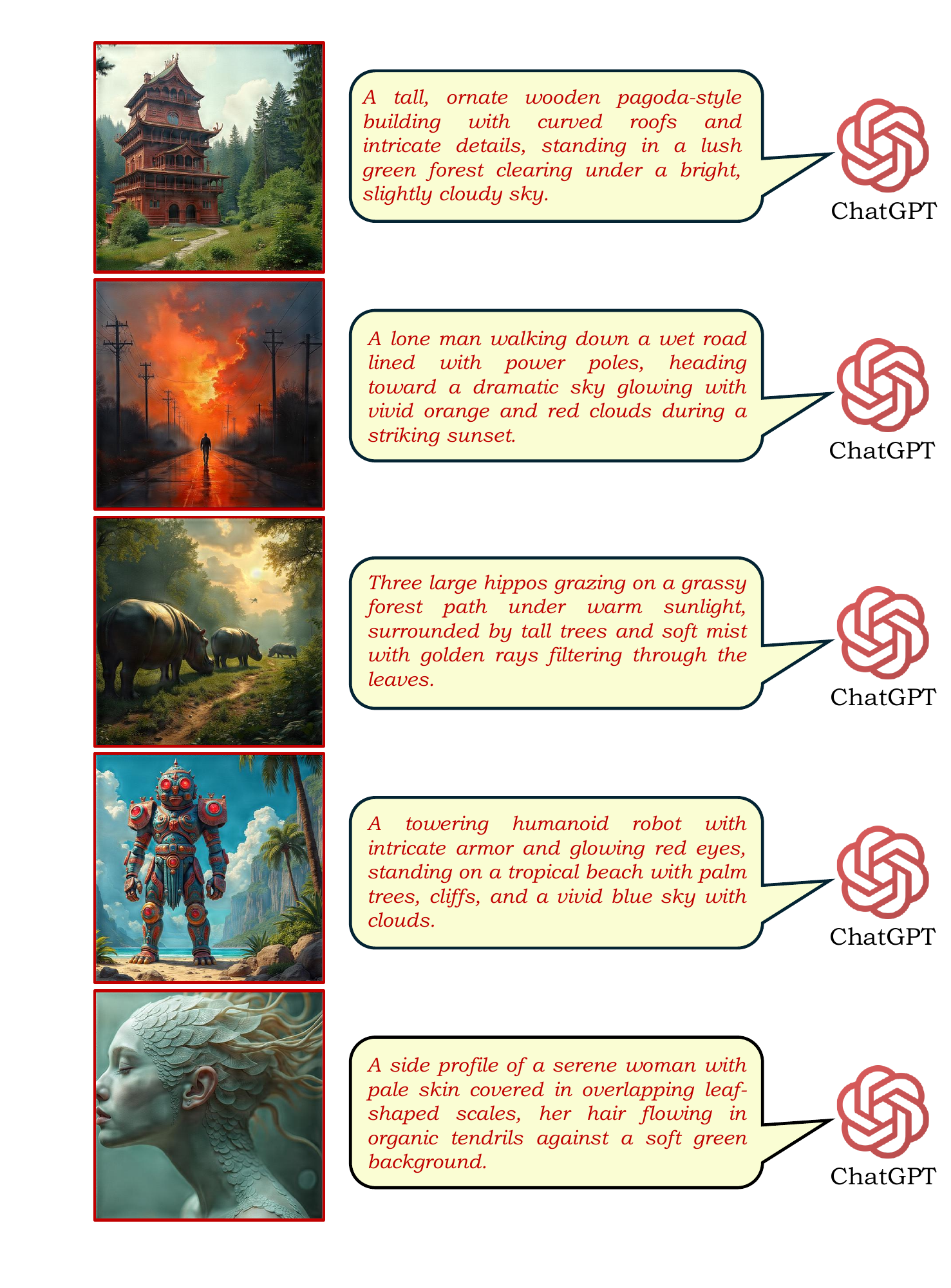}
    \vspace{-1em} 
    \caption{\small  Captions automatically generated by ChatGPT for a set of original watermarked images. These textual descriptions capture semantic and stylistic details of the inputs and are subsequently used to guide image-to-image diffusion for watermark removal.}
    \label{fig:diff_bbchat2}
\end{figure*}

\begin{figure*}[t]
    \centering
    \vspace{-2em} 
    \includegraphics[clip, trim=65 130 30 50, width=\textwidth]{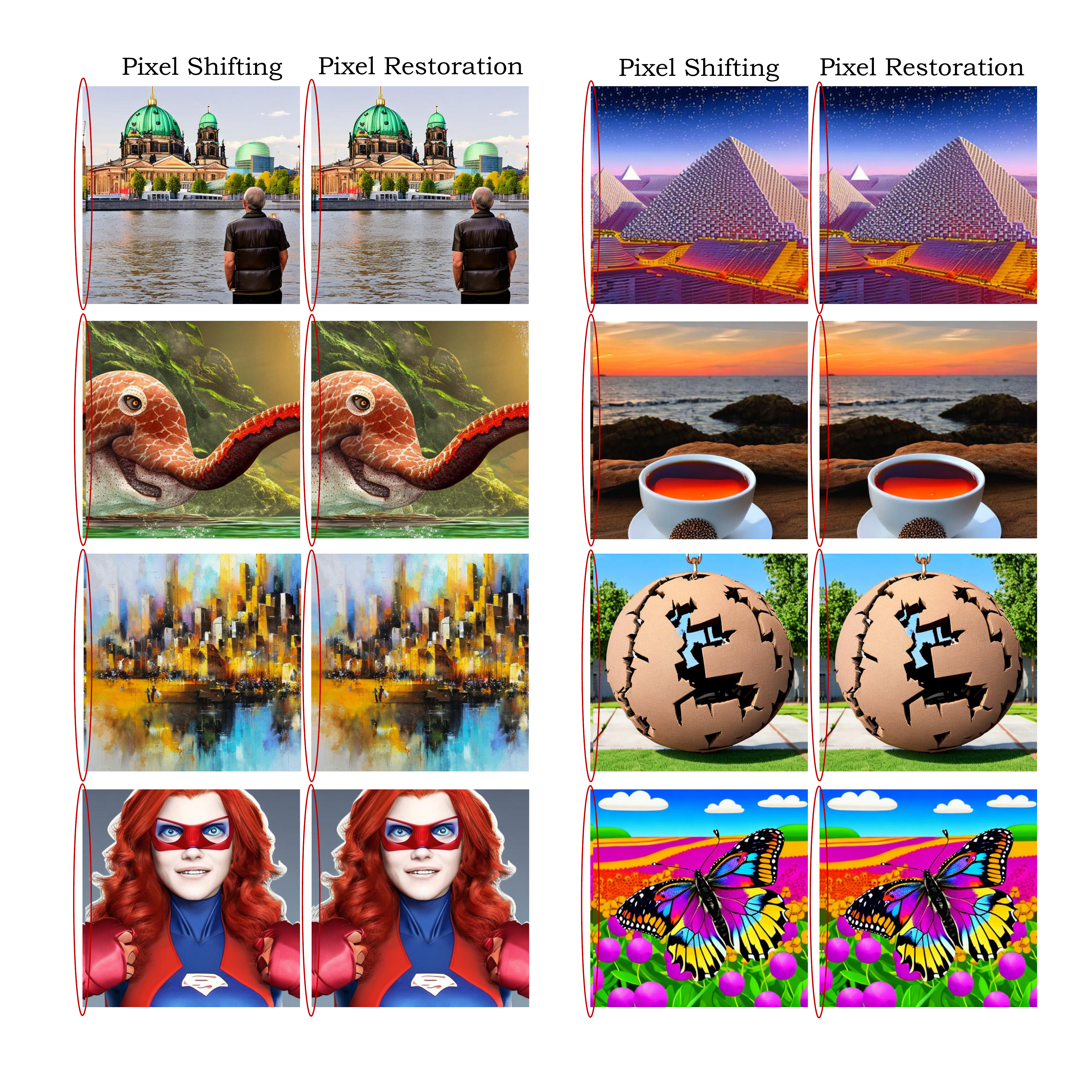}
    \vspace{-1em} 
    \caption{Effect of spatial translation on TreeRing watermarks. \textbf{Left}: Directly shifting the image by 7 pixels removes the watermark but introduces visible boundary artifacts. \textbf{Right}: Selectively restoring the shifted columns from the original image eliminates artifacts while preserving image quality, showing that minimal pixel-level restoration is sufficient.}
    \label{fig:diff_trans}
\end{figure*}

\begin{figure*}[t]
    \centering
    \vspace{-2em} 
    \includegraphics[clip, trim=65 45 10 25, width=\textwidth]{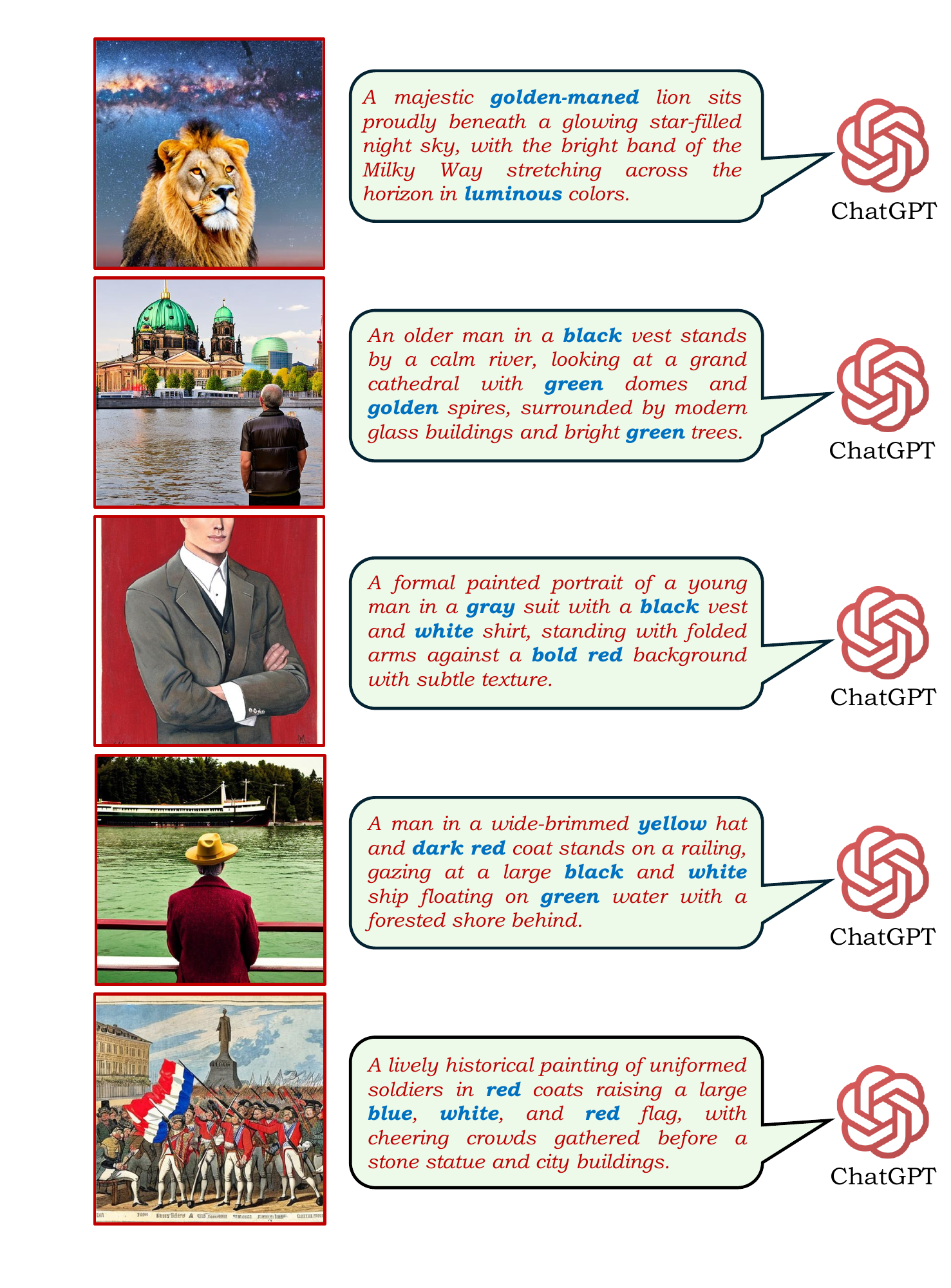}
    \vspace{-1em} 
    \caption{\small  Captions generated by ChatGPT for watermarked images, used in conjunction with a ControlNet-based image-to-image diffusion model. Unlike earlier setups, the prompts explicitly capture both semantic content and color attributes, providing richer conditioning for watermark-free synthesis.}
    \label{fig:chatgpt-controlnet}
\end{figure*}

\AppLine{\textbf{{\color{blue}Background}}}{}{sec:background}
\AppLine{\textbf{{\color{blue}Figure}}~\ref{fig:diff_stega}}{Qualitative results of Beige-box track on StegaStamp images}{fig:diff_stega}
\AppLine{\textbf{{\color{blue}Figure}}~\ref{fig:diff_treering}}{Qualitative results of Beige-box track on TreeRing images}{fig:diff_treering}
\AppLine{\textbf{{\color{blue}Figure}}~\ref{fig:diff_bbnoartifact}}{Qualitative results of Black-box track on images with no artifacts}{fig:diff_bbnoartifact}
\AppLine{\textbf{{\color{blue}Figure}}~\ref{fig:diff_bbboundary}}{Qualitative results of Black-box track with boundary artifacts images}{fig:diff_bbboundary}
\AppLine{\textbf{{\color{blue}Figure}}~\ref{fig:diff_bbcirc}}{Qualitative results of Black-box track with circular Fourier artifacts images}{fig:diff_bbcirc}
\AppLine{\textbf{{\color{blue}Figure}}~\ref{fig:diff_bbsquare}}{Qualitative results of Black-box track with square Fourier artifacts images}{fig:diff_bbsquare}
\AppLine{\textbf{{\color{blue}Figure}}~\ref{fig:diff_bbstrength}}{Qualitative results of Black-box track with varying diffusion strength}{fig:diff_bbstrength}
\AppLine{\textbf{{\color{blue}Figure}}~\ref{fig:diff_bbchat1}}{Captions generated by ChatGPT for watermarked images}{fig:diff_bbchat1}
\AppLine{\textbf{{\color{blue}Figure}}~\ref{fig:diff_bbchat2}}{Captions generated by ChatGPT  for watermarked images}{fig:diff_bbchat2}
\AppLine{\textbf{{\color{blue}Figure}}~\ref{fig:diff_trans}}{Effect of spatial translation on TreeRing watermarks}{fig:diff_bbchat1}
\AppLine{\textbf{{\color{blue}Table}}~\ref{tab:treeringresults}}{Quantitative results of our approach with and w/o pixel restoration}{tab:treeringresults}
\AppLine{\textbf{{\color{blue}Figure}}~\ref{fig:chatgpt-controlnet}}{Captions generated by ChatGPT  for ControlNet model}{fig:chatgpt-controlnet}

\clearpage
\section*{Background} \label{sec:background}

\subsection{StegaStamp: Invisible Hyperlinks in Photographs}

StegaStamp~\cite{tancik2020stegastamp} introduced one of the first end-to-end learned approaches for embedding invisible information into images that remain robust under real-world conditions. The key idea is to train an encoder–decoder system jointly while simulating the distortions introduced by printing and re-capturing photographs, such as perspective changes, blur, noise, color variation, and compression. The encoder embeds a short bitstring into the image with minimal perceptual difference, while the decoder is trained to recover the message reliably after these transformations. To further improve resilience, error-correcting codes are used to ensure accurate recovery even when parts of the image are degraded. Experiments demonstrated that StegaStamp achieves high decoding accuracy across different combinations of printers, displays, and cameras, while maintaining near-identical image quality. This work showed that imperceptibility and robustness can be achieved together, moving beyond earlier digital-only steganography methods and establishing a framework for watermarking that extends into practical physical-world settings.

In the context of the NeurIPS watermarking competition, StegaStamp is particularly relevant because the Beige-box track uses a variant of its algorithm to generate watermarked images. Understanding its design principles and robustness objectives provides a clear basis for developing effective removal strategies, since the competition task required eliminating such watermarks while maintaining image quality.

\subsection{Tree-Ring Watermarks: Fingerprints for Diffusion Images}

Tree-Ring watermarks~\cite{wen2024tree} addressed the challenge of watermarking in the context of diffusion-based generative models. The approach embeds circular “tree-ring” patterns in the Fourier phase of generated images, producing signals that are imperceptible in the spatial domain but remain stable under common transformations such as resizing, compression, cropping, and mild editing. By embedding during the image generation process itself, rather than applying marks after generation, the method ensures that the watermark is more deeply integrated and harder to remove. Detection is carried out in the frequency domain using statistical classifiers, enabling reliable identification of watermarked content at scale. This design makes tree-ring watermarks more robust than many prior spatial-domain techniques and provides a practical mechanism for verifying the provenance of AI-generated content. The work highlights the importance of designing watermarking strategies that align with the properties of modern generative models, establishing a foundation for authenticity and accountability in synthetic media.

For the competition, Tree-Ring Watermarks formed the basis of another watermarking method in the beigebox track. Their reliance on structured frequency-domain perturbations made removal more challenging, requiring participants to design attacks that disrupt the Fourier-phase patterns while preserving spatial fidelity. Understanding the strengths of tree-rings thus directly shaped the strategies needed for effective watermark removal.

\clearpage

\begin{table}[t]
\centering
\caption{Quantitative comparison of our Beige-box removal strategy for TreeRing watermarks with and without pixel restoration. Incorporating selective restoration improves both fidelity metrics (FID, CLIP-FID) and perceptual similarity (SSIM, LPIPS), demonstrating that boundary-aware restoration enhances watermark removal without compromising quality.}
\small
\setlength{\tabcolsep}{6pt}
\begin{tabular}{lcccccc}
    \toprule
    \rowcolor{blue!20}\textbf{Method} & \textbf{PSNR}$\uparrow$ & \textbf{SSIM}$\uparrow$ & \textbf{LPIPS}$\downarrow$ & \textbf{FID}$\downarrow$ & \textbf{NMI}$\uparrow$ & \textbf{CLIPFID}$\downarrow$ \\
    \midrule
    \rowcolor{blue!5} w/o Pixels Restoration & \textbf{14.878} & 0.428 & 0.133 & 15.773 & 0.212 & 3.199 \\
    \rowcolor{blue!10} \textbf{Ours} & 14.353 & \textbf{0.434} & \textbf{0.124} & \textbf{12.531} & \textbf{0.218} & \textbf{1.622} \\
    \bottomrule
\end{tabular}
\label{tab:treeringresults}
\end{table}

